\documentclass[10pt,twocolumn,letterpaper]{article}

\usepackage[pagenumbers]{cvpr}      %

\newcommand{\TODO}[1]{}

\def\labelset{gold}
\def\resdir{\labelset}
\newcommand{\handfig}[2][\linewidth]{\includegraphics[width=#1]{#2}}
\makeatletter
\newcommand{\restableat}[2]{{\let\ewb@table\table\def\table[##1]{\ewb@table[#1]}\input{\resdir/tables/#2}}}
\makeatother

\newcommand{\ntestdrawn}{10,000\xspace}
\newcommand{\ntestsent}{9,900\xspace}
\newcommand{\ntestscored}{9,761\xspace}

\newcommand{\silverbanner}{}

\makeatletter
\providecommand{\@LN@col}[1]{}
\providecommand{\@LN}[2]{}
\makeatother

\usepackage{adjustbox}

\newcommand{\ewn}{E-Waste-Bench\xspace}
\newcommand{\gap}[1]{\textbf{G#1}}
\providecommand{\checkmark}{\ensuremath{\surd}}  %

\newcommand{\cmark}{\textcolor{green!50!black}{\ensuremath{\checkmark}}}
\newcommand{\xmark}{\textcolor{red!80!black}{\ensuremath{\times}}}

\usepackage{microtype}

\definecolor{cvprblue}{rgb}{0.21,0.49,0.74}
\usepackage[pagebackref,breaklinks,colorlinks,allcolors=cvprblue]{hyperref}

\def\confName{}
\def\confYear{}

\title{\ewn: A Large-Scale Benchmark for Visual Triage of Data-Center Hardware}

\author{Nader Lobandi \quad Mohammad Mehdi Hosseini \quad Lombe Chileshe \quad Mohammad H. Mahoor$^{*}$\\
University of Denver\\
2199 S. University Blvd., Denver, CO 80208\\
{\tt\small \{nader.lobandi, mohammadmehdi.hosseini, lombe.chileshe\}@du.edu}}

\def\buildwhat{0}
\if\buildwhat1 \fi
\if\buildwhat2 \fi

\begin{document}
\twocolumn[{%
\renewcommand\twocolumn[1][]{#1}%
\maketitle
\begin{center}
  \includegraphics[width=\textwidth]{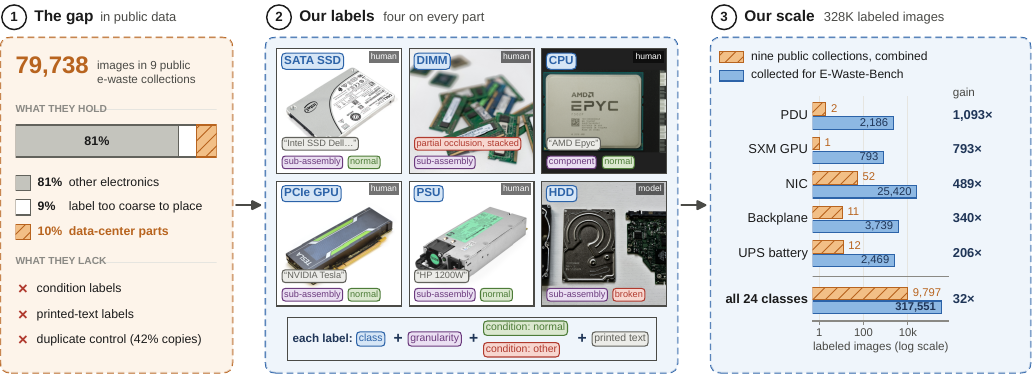}
  \captionof{figure}{What \ewn adds to public e-waste data. (1) By their own labels, only 10\% of the images in nine public collections show a data-center part, and none labels condition or printed text. (2) Every part carries a class (blue), a granularity (purple), a condition (green when normal, red otherwise) and its printed text (gray). The corner tag says who labeled a tile: annotators for test images, models for training images. (3) Collected images against public images of the same classes, counted by our class reading, for the five largest gains and for all 24 classes.}
  \label{fig:overview}
\end{center}
\vspace{1em}
}]
\iftoggle{cvprfinal}{{\let\thefootnote\relax\footnotetext{$^{*}$Corresponding author: {\tt mohammad.mahoor@du.edu}}}}{}
\silverbanner

\begin{abstract}
Retired data-center hardware can often be refurbished or resold, but triage demands more than object categories.
A model must tell visually similar parts apart, judge condition under damage and occlusion, and read the printed part number.
We audit nine public e-waste collections.
They cover mostly consumer devices (only 10\% of images are labeled as a data-center part), label neither condition nor printed text, and contain 42\% exact or augmented duplicates.
We introduce \ewn, 328K photographs drawn from 762K deduplicated images and labeled for class (24 classes), granularity, condition and printed text.
Images come from those collections and the web, where a recipe-based crawler works beside an agent built on an open-weight vision-language model.
The agent vets new sources and reads sites the crawler cannot parse, and both are steered toward under-represented classes.
Training labels are model-made, splits follow near-duplicate groups, and human annotators label a test set of nearly 10,000 images.
We benchmark classification, condition, detection, text perception and cross-dataset transfer.
On this test set, a ConvNeXt-T trained on \ewn reaches 64.3\% top-1, against 31.5\% with the public collections alone.
Near-duplicate leakage inflates top-1 by 3.0 points under model-made test labels but by less than 0.1 under human labels, so model-labeled benchmarks need group splits.
Accuracy falls to 42.1\% on marketplace and auction photographs, the closest proxy for hardware leaving service.
Data and code will be released.
\end{abstract}

\section{Introduction}
\label{sec:intro}

Data centers replace hardware constantly, and most of the drives, memory modules, power supplies, network cards and GPUs they retire are not waste yet.
They can be refurbished, redeployed, resold or mined for materials, at a scale that grows with generative AI~\cite{balde2024gem,wang2024genai} (supplementary \cref{sec:supp_motivation}).
Capturing that value starts with recognizing each part, and recognition must go beyond the object category.
A triage system has to tell an NVMe drive from a SATA drive that shares its form factor, and notice that a board is cracked, stacked or occluded.
It also has to read the part number that settles both identity and value (\cref{fig:overview}).

Existing datasets were not built for this task.
Waste-sorting benchmarks~\cite{bashkirova2022zerowaste,proenca2020taco} target materials and household litter, and the public e-waste collections~\cite{islam2023ewastenet,islam2024laptop} target consumer devices.
Auditing nine of them (\cref{sec:diagnosis}), we find four specific gaps.
\gap{1}: data-center hardware is nearly absent.
By the collections' own labels, only one image in ten shows a part of our 24-class taxonomy, and eight of the 24 classes have fewer than 100 images.
\gap{2}: no collection labels the condition a part is seen in.
\gap{3}: none pairs images with the text printed on the part.
\gap{4}: the collections are duplicated and noisy.
Within their own collection, 42\% of the images are exact or augmented copies, and half of the unique images recur in another collection.

Following AffectNet~\cite{mollahosseini2019affectnet,fard2025affectnetplus}, we engineer each step of \ewn against one of these gaps and measure its effect.
We gather photographs of data-center hardware from refurbisher catalogs, surplus auctions, public archives and manufacturer pages, and harmonize them with the public collections under one closed taxonomy (\gap{1}).
Every image is labeled on four axes that add condition (\gap{2}) and text (\gap{3}) to class and granularity.
Splits are assigned by near-duplicate group, with a test set of original photographs labeled by human annotators (\gap{4}).
Our contributions are
\begin{enumerate*}[label=(\arabic*)]
  \item a measured audit of nine public e-waste collections that puts a number on each gap,
  \item \ewn, 328,239 labeled photographs of data-center hardware with class, condition, granularity and printed-text labels, drawn from 762,129 unique images, with a test set labeled by human annotators and measured annotator agreement on every axis,
  \item a finding on leakage: near-duplicates of test images in training inflate top-1 far more under model-made test labels than under human labels, so model-labeled benchmarks need group splits as a control, and
  \item benchmarks for classification, condition, detection, text perception and cross-dataset transfer.
\end{enumerate*}

\section{Related work}
\label{sec:related}

Vision datasets for waste sorting have moved from single objects on plain backgrounds (TrashNet~\cite{yang2016trashnet}) to litter in the wild (TACO~\cite{proenca2020taco}), underwater debris~\cite{fulton2019trashicra} and cluttered indoor scenes~\cite{wang2020mjuwaste}.
ZeroWaste~\cite{bashkirova2022zerowaste} covers real material-recovery conveyors, and like its test data, ours holds original photographs only.
None of the four we checked annotates condition or printed text (supplementary \cref{tab:compare}).
Electronic-waste collections cover consumer devices and laptop parts~\cite{islam2023ewastenet,islam2024laptop,akshat2023ewaste}, shredded fractions~\cite{talwar2026aris}, printed circuit boards~\cite{soomro2022pcb,lu2020ficspcb} and community sets on Roboflow and Kaggle~\cite{cas2024pcparts,ewd2024r0ojc,manoj2024ewaste,ebs2025ewaste,asaniczka2024pcparts}.
Product recognition~\cite{goldman2019sku110k,wei2019rpc,bai2020products10k} is the closest work outside waste, but it assumes clean catalog or shelf imagery, whereas hardware leaving service is also photographed stacked, damaged and partly disassembled.

General recognition datasets~\cite{deng2009imagenet,lin2014coco,kuznetsova2020openimages} and long-tailed vocabularies~\cite{gupta2019lvis,shao2019objects365,vanhorn2018inaturalist} set the conventions we follow for taxonomy, splits and per-class reporting.
Web-scale collections with model-made labels or filters~\cite{schuhmann2022laion,gadre2023datacomp,kirillov2023sam} show that curation choices, not size alone, decide what a dataset teaches.
We do the same for training labels and keep human annotators for the test set, because two benchmark failures are well documented.
Near-duplicates across training and test inflate accuracy~\cite{barz2020cifair}, and re-collected test sets show how much fails to transfer~\cite{recht2019imagenet,barbu2019objectnet}.
Label errors pervade even standard test sets~\cite{northcutt2021pervasive,northcutt2021confident}.
We answer the first with group splits and the second with human test labels, and document the dataset with a datasheet~\cite{gebru2021datasheets}.
Scene-text benchmarks (TextVQA~\cite{singh2019textvqa}, TextOCR~\cite{singh2021textocr}) read text in the wild, and we pair hardware images with OCR readings~\cite{cui2025paddleocr}.
Open-vocabulary detectors (Grounding DINO~\cite{liu2023groundingdino}, OWLv2~\cite{minderer2023owlv2}) and promptable segmenters (SAM~\cite{kirillov2023sam}, SAM~3~\cite{carion2025sam3}) box objects without class-specific training.
We use them as box labelers and zero-shot baselines.

\section{What existing datasets miss}
\label{sec:diagnosis}

We audited the nine public e-waste collections we could obtain~\cite{islam2023ewastenet,islam2024laptop,akshat2023ewaste,cas2024pcparts,ewd2024r0ojc,manoj2024ewaste,ebs2025ewaste,asaniczka2024pcparts}, 87,830 images in total.
Two are versions of one project, distributed and used separately.
We assigned every image to our near-duplicate groups (\cref{sec:curation}) and mapped every native label by hand to one of our 24 classes, to \emph{other electronics} or to \emph{ambiguous}.
\Cref{tab:g4} sets them against \ewn.

\begin{table}[t]
\centering
\footnotesize
\setlength{\tabcolsep}{2.5pt}
\caption{Nine public e-waste collections against \ewn. Copies: exact or augmented duplicates inside a collection. Shared: images with a near-duplicate in another collection. DC: images labeled as one of our 24 classes. Last four columns: condition, printed text, granularity and object boxes annotated (\cmark) or not (\xmark). $^\dagger$Model-made boxes. $^\ddagger$Near-duplicates are kept and grouped, so none crosses a split.}
\label{tab:g4}
\begin{adjustbox}{max width=\linewidth}
\begin{tabular}{lrrrrcccc}
\toprule
collection & images & copies & shared & DC & \rotatebox{90}{condition} & \rotatebox{90}{text} & \rotatebox{90}{granularity} & \rotatebox{90}{boxes} \\
\midrule
Kaggle E-Waste~\cite{akshat2023ewaste} & 3,000 & 36 & 146 & 0 & \xmark & \xmark & \xmark & \xmark \\
PC Parts Det.~\cite{cas2024pcparts} & 978 & 570 & 0 & 701 & \xmark & \xmark & \xmark & \cmark \\
EBS E-Waste~\cite{ebs2025ewaste} & 4,627 & 3,248 & 505 & 346 & \xmark & \xmark & \xmark & \cmark \\
EWasteNet~\cite{islam2023ewastenet} & 881 & 15 & 28 & 0 & \xmark & \xmark & \xmark & \xmark \\
Kaggle PC Parts~\cite{asaniczka2024pcparts} & 3,279 & 203 & 32 & 957 & \xmark & \xmark & \xmark & \xmark \\
E-Waste Cls.~\cite{manoj2024ewaste} & 19,130 & 2,188 & 19,122 & 441 & \xmark & \xmark & \xmark & \xmark \\
Laptop Comp.~\cite{islam2024laptop} & 29,120 & 25,480 & 831 & 5,600 & \xmark & \xmark & \xmark & \xmark \\
WEEE, balanced~\cite{ewd2024r0ojc} & 7,211 & 1,278 & 6,880 & 380 & \xmark & \xmark & \xmark & \cmark \\
WEEE~\cite{ewd2024r0ojc} & 19,604 & 3,486 & 19,129 & 441 & \xmark & \xmark & \xmark & \cmark \\
\midrule
all nine & 87,830 & 36,504 & 39,437 & 8,866 & \xmark & \xmark & \xmark & 4/9 \\
\midrule
\textbf{\ewn} & \textbf{762K} & $\ddagger$ & $\ddagger$ & \textbf{328K} & \cmark & \cmark & \cmark & \cmark$^\dagger$ \\
\bottomrule
\end{tabular}
\end{adjustbox}
\end{table}

The first gap (\gap{1}) is the form factor.
By their own labels, 81\% of the images show a device that is not data-center hardware, and 9\% carry a label too coarse to place.
Only 8,866 (10\%) map to one of our 24 classes.
Coverage inside that tenth is uneven: in our labeled set, eight of the 24 classes have fewer than 100 public images (supplementary \cref{tab:classes}).
SXM GPU modules have one image, rack PDUs two and backplanes 11.
Two label axes are missing altogether (mark columns of \cref{tab:g4}).
No collection labels the condition a part is seen in (\gap{2}), yet damage, stacking and occlusion separate hardware leaving service from catalog photographs.
None pairs images with the text printed on the part (\gap{3}), but the model or part number often decides both the class and the resale or data-sanitization decision.

The last gap (\gap{4}) is duplication and label noise.
Inside a collection, 36,504 of the 87,830 images (42\%) are copies shipped as separate samples, 25,480 of 29,120 in the largest.
Across collections, 39,437 of the 79,738 unique images (49\%) have a near-duplicate in another one.
Nearly all are in three collections that are versions or re-uploads of one another.
So pooling them, or training on one and testing on another, tests on training photographs.
Where a native label maps to one of our classes (223 to 3,972 images per collection, supplementary \cref{tab:g4_full}), we compare it with an independent vision-language reading.
They agree on 71 to 90\% of images per collection, a disagreement and not an error rate, since that reading is model-made.

\section{The \ewn dataset}
\label{sec:dataset}

\begin{figure*}[t]
  \centering
  \handfig[\linewidth,trim=0 5 0 7,clip]{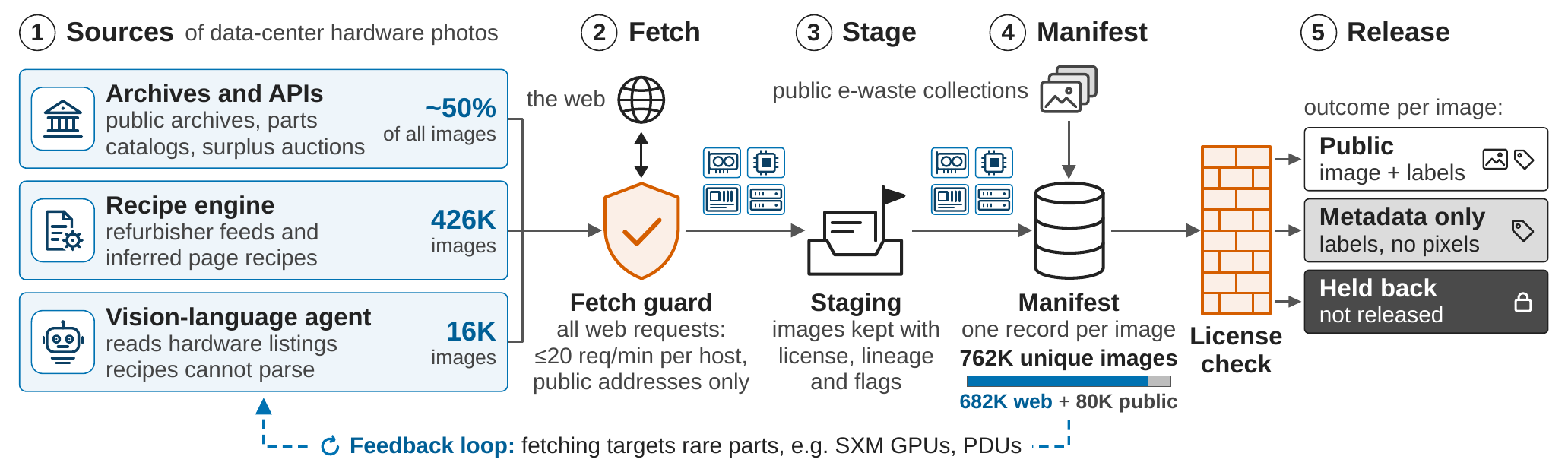}
  \caption{How the collected part of \ewn is built (\cref{sec:collection}). (1) Three lanes acquire images, steered by a feedback loop toward rare classes. (2) Every request passes one fetch guard. (3) Images are staged and flagged, never deleted. (4) A manifest records the provenance of every image. (5) An export firewall releases an image with its license evidence, its labels without the pixels, or nothing.}
  \label{fig:pipeline}
\end{figure*}

\Cref{fig:pipeline} summarizes how \ewn is built, and the subsections follow it.
Each step answers one gap of \cref{sec:diagnosis}, from the taxonomy (\gap{1}) and condition and text labels (\gap{2}, \gap{3}) to near-duplicate grouping and group splits (\gap{4}).

\subsection{Taxonomy and four label axes}
\label{sec:schema}
\ewn uses a closed set of 24 data-center hardware classes, plus a \emph{none} class for images that show no such part (\cref{fig:stats}a).
They span storage (hard drives, SATA and NVMe SSDs, drive caddies, backplanes, RAID cards and their batteries) and compute (CPUs, DIMMs, motherboards, PCIe and SXM GPUs, heatsinks, risers).
The rest are power (power supplies, UPS batteries, PDUs), networking (NICs, optical transceivers, DAC cables, top-of-rack switches) and enclosures (chassis, GPU server nodes, fans).
The classes are field-replaceable units that fleet studies document as failing or replaced in service~\cite{vishwanath2010characterizing,wang2017fouryears,kokolis2025reliability}.
They also cover parts handled in the same swaps and high-value parts recovered at decommissioning~\cite{wang2024greensku} (references per component in supplementary \cref{sec:supp_classes}).
The set is closed by design: a source category that does not map to one class leaves the image unclassified.
Each image has four label axes.
The \emph{class} is one of the 25 labels above.
The \emph{condition} is a multi-label set covering damage (broken, deformed) and the state a part is photographed in (stacked, partly occluded, unusual angle), with \emph{normal} when none applies (\gap{2}).
The \emph{granularity} is a whole device, a sub-assembly that installs into a device, or a single component.
The \emph{text} is the lines printed on the part with their type, model or part number, capacity or certification mark (\gap{3}).

\subsection{Collection}
\label{sec:collection}
Two pipelines feed one manifest (\cref{fig:pipeline}).
The curated pipeline ingests the nine public collections with their native labels kept as provenance.
The collection pipeline gathers photographs of new, used and decommissioned data-center hardware from 150 web sources, 143 of which supply labeled images.
The 131 that it discovered on its own supply 277,085 of the labeled images.
The rest are 4 public archives, 4 refurbisher catalogs, a manufacturer, a product catalog, an auction site and a forum.
New sources come from a seed list, a web-search API and links on pages already read.
An open-weight vision-language model, Qwen3.5-35B-A3B~\cite{qwen2026qwen35}, served locally on one GPU, judges each candidate site.

A feedback loop from the manifest steers acquisition toward rare parts.
Each class has a target count, and its deficit, recomputed from the manifest between passes, ranks the next jobs.
The four classes that no public collection labels (drive caddies, RAID batteries, optical transceivers, UPS batteries) come first in search queries and weigh most in ranking a site's listing pages.
A source is then read, in order of preference, through an official API, by the recipe engine, or by the vision-language agent.
The recipe engine reads a store's product feed where one exists.
Otherwise it infers a site's link templates, image pattern, full-size rewrite and pagination as declarative data from sampled listing and product pages.
It verifies them through the production extractor, then walks the site with no model in the loop, executing nothing derived from a page.
It yields 425,709 unique images, 62\% of those collected.
The vision-language agent, the same model run as four parallel workers, takes the listings the recipe engine cannot parse.
It browses with a single page-fetch tool for at most 24 rounds per job and labels each image it finds.
Its output is validated as untrusted input, and an image URL that never appeared on a page it fetched is dropped. The agent adds 16,161 unique images.
Every request passes a fetch guard that holds each host to 20 requests per minute across all processes and checks that each redirect hop resolves to a public address.
Yield feeds back into scheduling: a host that stops returning new images is set aside, and an exhausted one is revisited for new listings.

Of the 328,239 labeled images, 318,442 were collected, and the 9,797 public ones enter training only.
Validation and test hold collected images only.
Access rules, rate limits and the licensing review are in supplementary \cref{sec:supp_collection}.

\subsection{Curation}
\label{sec:curation}
\begin{figure}[!b]
  \centering
  \includegraphics[width=\linewidth,trim=0 5 0 5,clip]{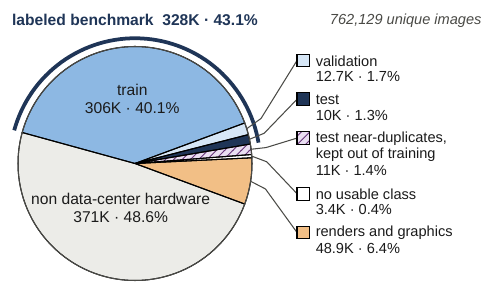}
  \caption{Composition of the 762,129 unique images. The arc marks the labeled benchmark (train, validation, test), 43.1\% of the corpus. Near-duplicates of test images are kept out of training, and images that show no data-center hardware are kept as the class \emph{none}. Exact counts are in supplementary \cref{tab:funnel}.}
  \label{fig:composition}
\end{figure}
Exact duplicates are removed by content hash, leaving 762,129 unique images: 682K of the 1.45 million fetched from the web and 80K public ones (\cref{fig:composition}, exact counts in supplementary \cref{tab:funnel}).
Near-duplicates, such as the same photograph resized, recompressed, watermarked by a reseller or augmented, are grouped by a 64-bit perceptual hash without chaining matches (\cref{eq:group}).
An earlier transitive clustering merged 76,558 white-background product shots into one group.
Grouping yields 389,169 groups, 70\% of them single images, with the largest holding 943 (supplementary \cref{fig:supp_stats}b).
Near-duplicates are kept and flagged, never dropped: a resold copy has its own watermark and framing.

Auction listings and archives also hold furniture, vehicles and tools, and product pages hold renders, diagrams and logos.
A vision-language model therefore sorts each image into hardware, another object, or a non-photograph, and near-duplicates of a non-hardware image inherit its reading.
This leaves 328,239 labeled hardware images, the benchmark proper.
Another 370,665 non-hardware images, photographs that show no data-center part, are kept as the \emph{none} class in training.
Renders and graphics (48,866) only train a relevance filter for later collection.
Near-duplicates of test images (10,955) are withheld from every split, and 3,404 images without a usable class are unused.

\subsection{Labels}
\label{sec:labels}
Training and validation labels are pseudo-labels made by models.
The same vision-language model, Qwen3.5-35B-A3B~\cite{qwen2026qwen35}, reads each image against written definitions of the 24 classes and gives its class, condition and granularity.
An image the model could not place falls back to its source category or mapped native label.
SAM~3~\cite{carion2025sam3} and OWLv2~\cite{minderer2023owlv2}, prompted once per class, label the objects with outlines and boxes that carry the prompted class.
Boxes of any class at IoU 0.7 or more are merged into one object with a single class.
Text is read by an OCR model, PP-OCRv5~\cite{cui2025paddleocr}, and by the vision-language model.
About one image in nine carries a reseller overlay (a shop name or web address printed over the photograph).

\begin{figure}[t]
  \centering
  \includegraphics[width=\linewidth,trim=0 17 0 12,clip]{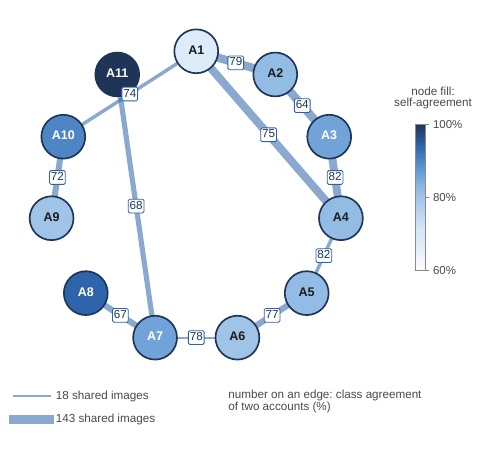}
  \caption{Agreement on the test images answered twice. An edge joins two accounts that answered the same images, labeled with their class agreement on final answers (\%). Node fill is an account's agreement with itself on first answers to repeats. Design: supplementary \cref{fig:annotation_design}.}
  \label{fig:annotation_overlap}
\end{figure}
Human annotators label the test set.
We draw \ntestdrawn collected photographs, stratified by pseudo-label class, from a pool of one image per near-duplicate group and a short side of at least 512 pixels.
Up to a quarter of each class deliberately comes from images where the pseudo-labelers disagree with each other or the source category.
We set 100 aside as controls and send the other \ntestsent to annotators.

Eight annotators took on the eleven annotation jobs, each under its own account.
They see each image with its main object outlined and three class suggestions, and confirm or correct its class, condition and granularity.
They may answer \emph{none} outside the 24 classes, and type the identifying text printed on it or mark its absence.
Part is answered twice, blind by a second annotator or by the same one about 200 tasks later, 11,457 tasks in all (\cref{fig:annotation_overlap}).
An image's class is the majority of its confident answers.
Every score uses the \ntestscored images whose answers settle on one class (ties in supplementary \cref{sec:supp_annotation}).

Two human annotators choose the same class on 74.0\% of 1,057 doubly answered images (Cohen's $\kappa = 0.72$).
One annotator repeats their own class on 84.0\% of 500 (per account and per axis in supplementary \cref{fig:annotation_agreement,tab:agreement}).
More than half of the disagreement is about \emph{none}.
In 15\% of pairs one annotator names a part and the other answers \emph{none}, against 11\% where they name two different parts.
The pseudo-label matches the human class on 77.9\% of the test images annotators assigned to one of the 24 classes.
It gave a class to four in five of the 21\% that they marked \emph{none}.
The test draw over-samples disputed images, so these figures are pessimistic for the training set.
They still show noisy training labels, with \emph{none} their weakest point.

\subsection{Splits and leakage prevention}
\label{sec:splits}
Every image $x$ belongs to one near-duplicate group $g(x)$.
Images are visited originals before augmented copies and larger before smaller, and an image not yet grouped becomes a center $c$ that takes every ungrouped image $x$ with
\begin{equation}
  d_H\big(h(x), h(c)\big) \le 6,
  \label{eq:group}
\end{equation}
where $h$ is the 64-bit perceptual hash and $d_H$ the Hamming distance, so group mates differ from their center in at most 6 of 64 bits.
An augmented copy shipped by a public collection joins the group of its source image unless that group would pass 500 images.
Splits are assigned by group, and validation is 5\% of the remaining groups, stratified by class and used for model selection only.
For a training set $\mathcal{T}$ with groups $G_{\mathcal{T}} = \{g(y) : y \in \mathcal{T}\}$ and a test set $\mathcal{S}$, the leakage rate is the share of test images with a group mate in training,
\begin{equation}
  L(\mathcal{T}, \mathcal{S}) = \frac{\big|\{x \in \mathcal{S} : g(x) \in G_{\mathcal{T}}\}\big|}{|\mathcal{S}|},
  \label{eq:leak}
\end{equation}
which is 0 for our split by construction and 0.75 for a random image split.
Of our test images, 37\% have a near-duplicate that the group split withholds, which \cref{sec:abl} uses to measure what leakage is worth.

\subsection{Statistics}
\label{sec:stats}
\begin{figure*}[t]
  \centering
  \includegraphics[width=\linewidth]{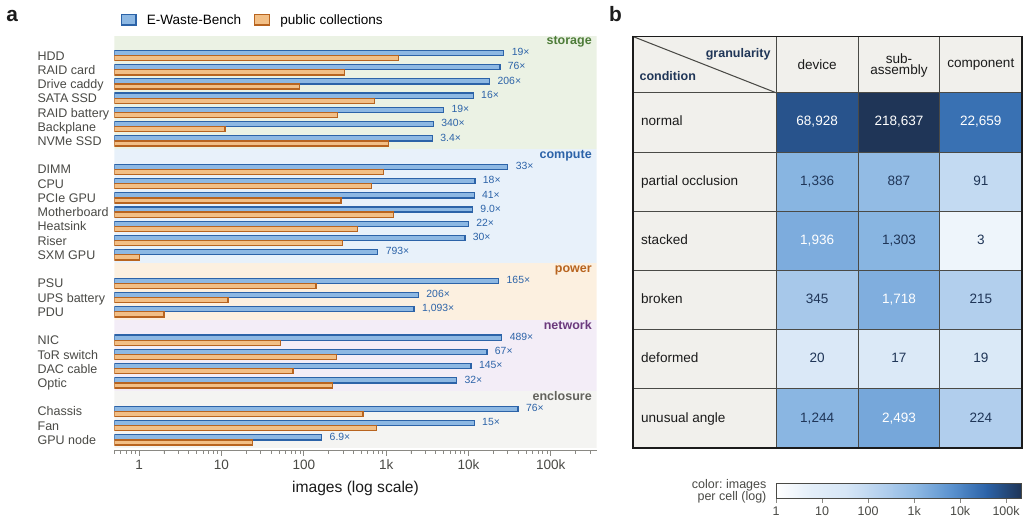}
  \caption{Statistics of the labeled images. (a) Labeled images per class in \ewn and in the nine public collections (log axis, grouped by function), with our gain over public data beside each class. (b) Images per condition (rows) and granularity (columns). An image with two conditions counts in both rows, and shading follows the count on a log scale. Outside the test set, all labels are model-made.}
  \label{fig:stats}
\end{figure*}
\Cref{fig:stats} summarizes the labeled images (counts per class and split in supplementary \cref{tab:classes}).
Twenty-two of the 24 classes have more than 2,000 collected images (a, \gap{1}).
Per class, \ewn holds 3.4 to 1,093 times as many images, 41 times at the median.
Still, SXM GPU modules (793) and GPU server nodes (165) remain rare.
Panel (b) shows how rarely pseudo-labels mark a condition other than normal, deformation on 56 images (\gap{2}).

\section{Benchmarks}
\label{sec:bench}

\subsection{Protocol}
\label{sec:protocol}
All benchmarks use the frozen release snapshot, its group splits and the \ntestscored scored test images (\cref{sec:labels}).
Annotators assigned each to one of the 24 classes (7,700) or to \emph{none} (2,061, 21\%).
We report top-1 on all images, top-1 on the images the annotators \emph{named} and the recall of \emph{none}.
The test set is balanced by class and training is long-tailed, so macro-F1 and accuracy on head, mid and tail classes (by training frequency) are also reported.
Models are selected on validation and tested once, and classification results are mean and standard deviation over three seeds.
Each condition value is scored by average precision, whose chance level is the value's share.
Annotators labeled the class of the outlined main object and checked no box, so detection reports whether a detector finds that object and gives it the human class.

The classifiers are four fine-tuned families trained with one recipe: a standard reference (ResNet-50~\cite{he2016resnet}), a modern convolutional network (ConvNeXt-T and -B~\cite{liu2022convnext}), a vision transformer (ViT-B/16~\cite{dosovitskiy2021vit}) and an efficiency-oriented network (EfficientNet-B3~\cite{tan2019efficientnet}).
Probes on frozen self-supervised (DINOv2~\cite{oquab2024dinov2}) and language-image (CLIP~\cite{radford2021clip}) features join them, with CLIP zero-shot as the floor.
Gemma~3 27B~\cite{gemmateam2025gemma3} runs zero-shot too: it made none of our labels, and no annotator saw its answers.
The three trained detectors share one codebase, backbone and 12-epoch schedule.
YOLO11-m~\cite{jocher2024yolo11} is the real-time reference, and two open-vocabulary detectors run zero-shot.
Qwen3.5, SAM~3 and OWLv2 made the pseudo-labels (\cref{sec:labels}) and the suggestions annotators saw, so their rows are marked (settings in supplementary \cref{sec:supp_hparams}).

\subsection{Device classification}
\label{sec:cls}
\begin{figure*}[t]
  \centering
  \includegraphics[width=0.95\linewidth]{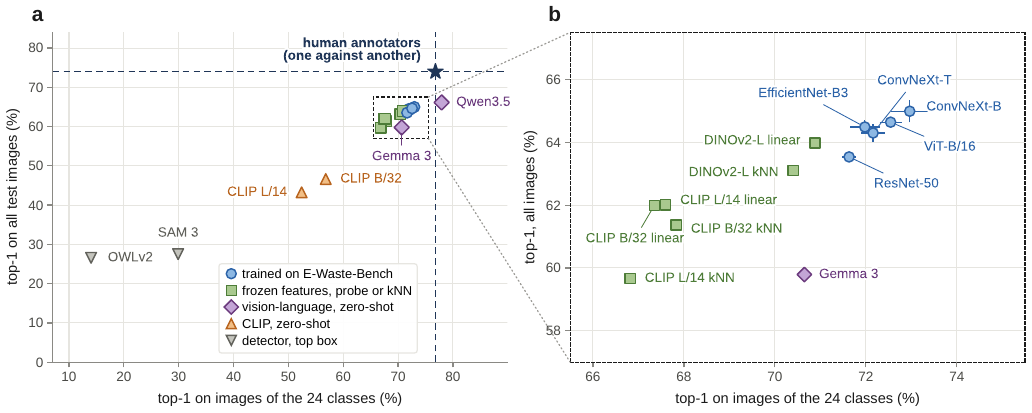}
  \caption{One point per classifier: top-1 on the images annotators gave one of the 24 classes (\emph{named}, x) against all scored test images (y), from \cref{tab:cls}. (a) All models by family. The star scores one annotator against another on the 1,057 doubly answered images. (b) The dashed box of (a), enlarged and labeled. Bars: standard deviation over three seeds, probes are one run.}
  \label{fig:cls_scatter}
\end{figure*}

\restableat{!t}{classification}
In \cref{tab:cls} and \cref{fig:cls_scatter}, the five fine-tuned models land within 1.5 points (63.5 to 65.0\% top-1, ConvNeXt-B best).
A linear probe on frozen DINOv2-L reaches 64.0\%, suggesting that training-label noise, not the backbone, sets the ceiling.
CLIP zero-shot stays at 43 to 47\%.

Two vision-language models run zero-shot with the same prompt, class definitions, image size and decoding.
Qwen3.5, which also produced the training labels, reaches 66.2\% top-1, the highest in the table, and Gemma~3 27B, which took no part in building the dataset, 59.8\%.
One annotator agrees with another on 74.0\% (\cref{sec:labels}).
On the 782 images where two different annotators agree, the best trained classifier reaches 77.1 $\pm$ 1.0\% (ConvNeXt-B).
About a quarter of these unambiguous images are thus misclassified, so ambiguous test labels do not explain the errors.

Three things remain challenging.
First, \emph{none}: of the 2,061 images annotators marked \emph{none}, the trained classifiers recover only 33 to 37\%.
Hardware outside the 24 classes is the larger and harder part of them (supplementary \cref{fig:none_split}).
Second, the tail: ConvNeXt-B reaches 57.9\% on tail classes (\cref{tab:cls}) against 77.9\% on mid-frequency ones (supplementary \cref{tab:cls_full}).
Third, the rarest classes: GPU server nodes (2.3\% correct of 57 test images) and SXM GPU modules (11.6\% of 49) are not learned from 109 and 604 training images.
The pseudo-label itself matches the annotators on only 31 of the 57 nodes.
Next come backplanes (32.7\%), PDUs (38.6\%) and UPS batteries (43.4\%), against 86\% or more for CPUs, PCIe GPUs, DIMMs and hard drives (supplementary \cref{fig:confusion,fig:failures}).

Slices of the test set (supplementary \cref{tab:slices_full}) find ConvNeXt-T weakest on the 178 marketplace and auction photographs, the closest we have to hardware as it leaves service.
There it reaches 42.1\% top-1, against 84.7\% on the 297 product-catalog shots.
No source is the floor of a recycling facility.
Images drawn on purpose where the pseudo-labelers disagreed are 7 points harder (58.9 against 65.9\%), so the test set is not limited to what those models get right.
A reseller overlay costs 2.1 points.

\subsection{Condition and granularity}
\label{sec:cond}
A multi-task head on ConvNeXt-T and on ViT-B/16 predicts condition and granularity alongside the class, with scores below in that order (supplementary \cref{tab:cond_full}).
Condition is the weakest axis for models and annotators alike, with noisy labels on both sides.
We report normal against not normal, stacked, and damaged, pooling the rare values broken and deformed (91 images).
All six values stay in the released labels.
Only stacking reaches moderate agreement (Krippendorff's $\alpha$ 0.50).
It is also the one value models find far above chance (AP 44.4 and 46.6 against 3.4\%).
Damage is rare and annotators do not agree on it ($\alpha$ 0.08 for the pooled value, 0.11 for broken alone), so pooling does not make it more reliable.
Models reach only AP 10.3 and 7.2.
Normal against not normal reaches AP 87.9 and 88.3, little above its 82.1\% share, with $\alpha$ 0.23, and granularity accuracy is 66\%.
The training pseudo-labels mark under 4\% of images as other than normal, where annotators mark 18\% of the test images.
Class accuracy falls from 67\% on images they call normal to 56\% under partial occlusion and 52\% from an unusual angle (supplementary \cref{fig:supp_cond}), the practical cost of \gap{2}.

\subsection{Detection}
\label{sec:det}
\restableat{!b}{detection}
We train Faster R-CNN~\cite{ren2015faster}, RetinaNet~\cite{lin2017focal}, FCOS~\cite{tian2019fcos} and YOLO11-m on the pseudo-labeled boxes, and prompt Grounding DINO~\cite{liu2023groundingdino} and OWLv2~\cite{minderer2023owlv2} zero-shot with one everyday name per class, chosen on validation data (\cref{tab:det}).
When one of its detections (up to 100 per image) overlaps the outline box at $\mathrm{IoU} \ge 0.5$, a detector \emph{finds} the outlined object.
It \emph{names} the object when the highest-scoring of those detections has the human class (supplementary \cref{eq:det}).
Both are means over the 7,512 test images with a hardware class from annotators and an outline box.
Finding the part is nearly solved and naming it is not.
The three torchvision detectors find the outlined object on 94 to 97\% of images and name it on 59 to 60\%.
YOLO11-m is the best trained detector on both (97.5 and 63.4\%).
Grounding DINO and OWLv2 zero-shot find it on 99\% of images but name it on only 13 and 25\%.
The vocabulary of data-center parts thus lies outside what web-scale grounding data covers.
A detector trained only on boxes both pseudo-labelers agreed on names 31\%, half of what training on all boxes gives.
Detection quality thus depends directly on the box labels (COCO AP in supplementary \cref{tab:det_full}).

\subsection{Text perception}
\label{sec:text}
Over half of the labeled images carry text (54.3\%), and 29\% show a model or part number.
The OCR and vision-language readers agree exactly on only 47.2\% of lines, with a median character error rate of 0.161 between them (supplementary \cref{fig:text_stats,tab:text}).
Where annotators typed an identifier, the OCR reader's output contains it on 35.3\% (of the 2,950 it returned text for) and the vision-language reader's on 41.3\% (of 2,729).
On the 2,729 both read, 55.6\% contain it in neither (supplementary \cref{fig:text}), so reading identifiers on hardware photographs is far from solved.
The text is nonetheless useful evidence.
On the 6,366 scored test images that carry text, a linear probe on frozen CLIP ViT-L/14 image features reaches 60.9\% top-1.
Adding the embedded OCR text raises it to 62.3\% (one seed), a small gain that a better reader should enlarge.

\subsection{Cross-dataset transfer}
\label{sec:cross}
\begin{figure}[t]
  \centering
  \includegraphics[width=\linewidth]{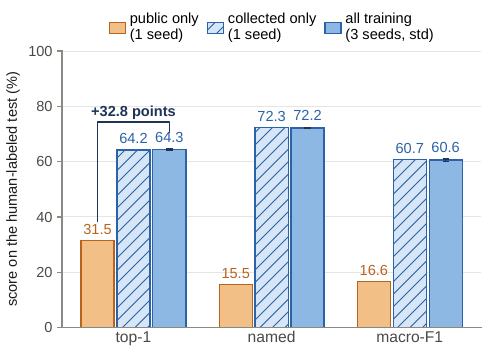}
  \caption{ConvNeXt-T trained on each set and tested on the human-labeled test images. Training on \ewn adds 32.8 top-1 points over the public collections alone (32.7 with its collected images only), and adding the public images changes almost nothing. Only the all-training bars average three seeds.}
  \label{fig:cross}
\end{figure}
Training ConvNeXt-T on the public collections, on our collected images, and on both measures the cost of \gap{1} directly (\cref{fig:cross}, supplementary \cref{tab:cross}).
The public collections give 31.5\% top-1, and only by answering \emph{none}, since the model is right on 15.5\% of the images annotators named.
The collected images give 64.2\% (72.3\% on named images), and adding the public collections changes almost nothing (64.3\%).
Conversely, the model trained on collected images reaches 59.1\% on 2,607 public images under their own labels, in the eight classes those labels cover.

\subsection{Leakage, ablations and scaling}
\label{sec:abl}
With the test set fixed, we put the withheld near-duplicates of test images back into training, which lifts the leakage rate of \cref{eq:leak} from 0 to 0.37.
On human labels, top-1 rises by less than 0.1 points, inside the 95\% interval even on the 3,588 test images that have such a duplicate (supplementary \cref{tab:slices_full}).
Scored against pseudo-labels, the same change gains 3.0 points, because a memorized duplicate reproduces its group's pseudo-label, right or wrong.
Group splits therefore matter most where test labels are model-made.
Of the other ablations (supplementary \cref{sec:supp_abl}), only dropping the non-hardware images from training costs more than a point (6.4, since a fifth of the test images are \emph{none}).
Removing training images that share a source page with a test image, which groups do not catch, costs 0.7.
Accuracy rises from 56.9\% with 5\% of the training groups to 64.3\% with all and flattens after 25\%.
But scores on model-made labels keep rising (supplementary \cref{fig:scaling,fig:supp_silvergold}).
Cleaner labels, not more of the same data, are therefore the next step.

\section{Privacy, licensing and ethics}
\label{sec:responsible}

\ewn depicts hardware and holds no personal profiles or account data, but people appear incidentally.
A privacy audit with SAM~3~\cite{carion2025sam3} flagged 3,557 of the 328,239 usable images (1.1\%), and released copies blur the detected faces and plates.
A manual check of the 404 flagged test images gives a per-image blur precision of 94.4\% and recall of 91.8\%.
The annotators were paid graduate students in electrical and computer engineering, familiar with data-center hardware and trained with a tutorial and a detailed manual.
Our institution's review board examined the annotation protocol and determined that it is not human-subjects research.
Every image keeps its source URL, collection time and the source policy in force.
Licensing is recorded per image and reviewed after collection, and images whose license does not permit redistribution are excluded at release export.
Documented removal requests are honored through a logged process that also blocks the source.
Supplementary \cref{sec:supp_ethics} gives the full statement, including derived images and misuse.
The annotations are released under CC BY-NC 4.0, with terms of use that restrict the data to research.

\section{Conclusion}
\label{sec:conclusion}

\ewn is, to our knowledge, the largest image corpus of data-center hardware: 328K labeled photographs against fewer than 10K in nine public collections.
It lifts top-1 on its human-labeled test set from 31 to 64\%.
What remains is rejecting images outside the 24 classes, the rarest classes, condition, hardware as it leaves service and reading printed identifiers.
For these, human labels show that cleaner training labels come next.

\section*{LLM Usage Statement}

Large language models were used in this work in three ways.
First, as part of the method: the open-weight vision-language model Qwen3.5-35B-A3B~\cite{qwen2026qwen35}, served locally, judged candidate image sources, browsed sites that the recipe engine could not read, and is one of the pseudo-labelers for class, condition, granularity and text (\cref{sec:collection,sec:labels}). Its class suggestions were shown to the human annotators, and it is evaluated zero-shot beside Gemma~3~\cite{gemmateam2025gemma3}.
Second, an LLM-based coding assistant (Claude, Anthropic) wrote and tested most of the collection, curation, benchmark and plotting code under the authors' direction and review, and every reported number, table and generated figure was produced by that reviewed code.
Third, the same assistant drafted and edited parts of the text, which the authors revised, proposed related references, and helped assemble the bibliography, every entry of which was checked against a publisher or archive record.
The authors designed the study, made every methodological decision, and take full responsibility for the content.

{
    \small
    \bibliographystyle{ieeenat_fullname}
    \bibliography{main}
}

\clearpage

\renewcommand{\floatpagefraction}{0.7}
\renewcommand{\dblfloatpagefraction}{0.82}
\setcounter{dbltopnumber}{3}
\renewcommand{\dbltopfraction}{0.85}
\makeatletter
\setlength{\@fptop}{0pt}
\setlength{\@dblfptop}{0pt}
\setlength{\@fpsep}{12pt plus 2pt}
\setlength{\@dblfpsep}{12pt plus 2pt}
\makeatother
\maketitlesupplementary
\appendix

\newcommand{\tocgroup}[1]{\par\addvspace{4pt}\noindent\textbf{#1}\par}
\newcommand{\tocitem}[2]{\par\noindent\hspace*{1em}\hyperref[#1]{\ref*{#1}.~#2}\nobreak\dotfill\hyperref[#1]{\pageref*{#1}}\par}
\noindent{\large\bfseries Contents}\par\smallskip
{\small
\tocgroup{Motivation and future work}
\tocitem{sec:supp_motivation}{Why data-center hardware}
\tocitem{sec:supp_future}{Future work}
\tocgroup{The dataset}
\tocitem{sec:supp_gallery}{Class gallery}
\tocitem{sec:supp_datasheet}{Datasheet}
\tocitem{sec:supp_classes}{Images per class}
\tocitem{sec:supp_collection}{Collection and source policy}
\tocitem{sec:supp_audit}{Full audit of the public collections}
\tocitem{sec:supp_waste}{Comparison with general waste datasets}
\tocitem{sec:supp_figs}{More statistics}
\tocgroup{Annotation and labels}
\tocitem{sec:supp_annotation}{Annotation protocol}
\tocitem{sec:supp_labelers}{Every source of labels against people}
\tocitem{sec:supp_silvergold}{Model-made against human labels}
\tocitem{sec:supp_anchoring}{Did the suggestions steer the annotators?}
\tocitem{sec:supp_class_reliability}{Per-class reliability of the test set}
\tocitem{sec:supp_foldft}{Model-made against human labels in training}
\tocgroup{Benchmarks}
\tocitem{sec:supp_cls}{More on classification errors}
\tocitem{sec:supp_cond}{Condition values one by one}
\tocitem{sec:supp_det}{Detection with COCO AP}
\tocitem{sec:supp_det_topk}{Detection with fewer boxes per image}
\tocitem{sec:supp_abl}{Ablations and scaling}
\tocitem{sec:supp_hparams}{Training settings and software}
\tocitem{sec:supp_tables}{Tables behind the figures}
\tocitem{sec:supp_source_exposure}{Accuracy and the training exposure of a source}
\tocgroup{Responsible use}
\tocitem{sec:supp_ethics}{Privacy, licensing and ethics in full}
\tocitem{sec:supp_privacy}{Privacy examples}
\tocitem{sec:supp_credits}{Image credits}
\par}
\medskip

\medskip

\section{Class gallery}
\label{sec:supp_gallery}
\begin{figure*}[tp]
  \centering
  \includegraphics[width=\linewidth]{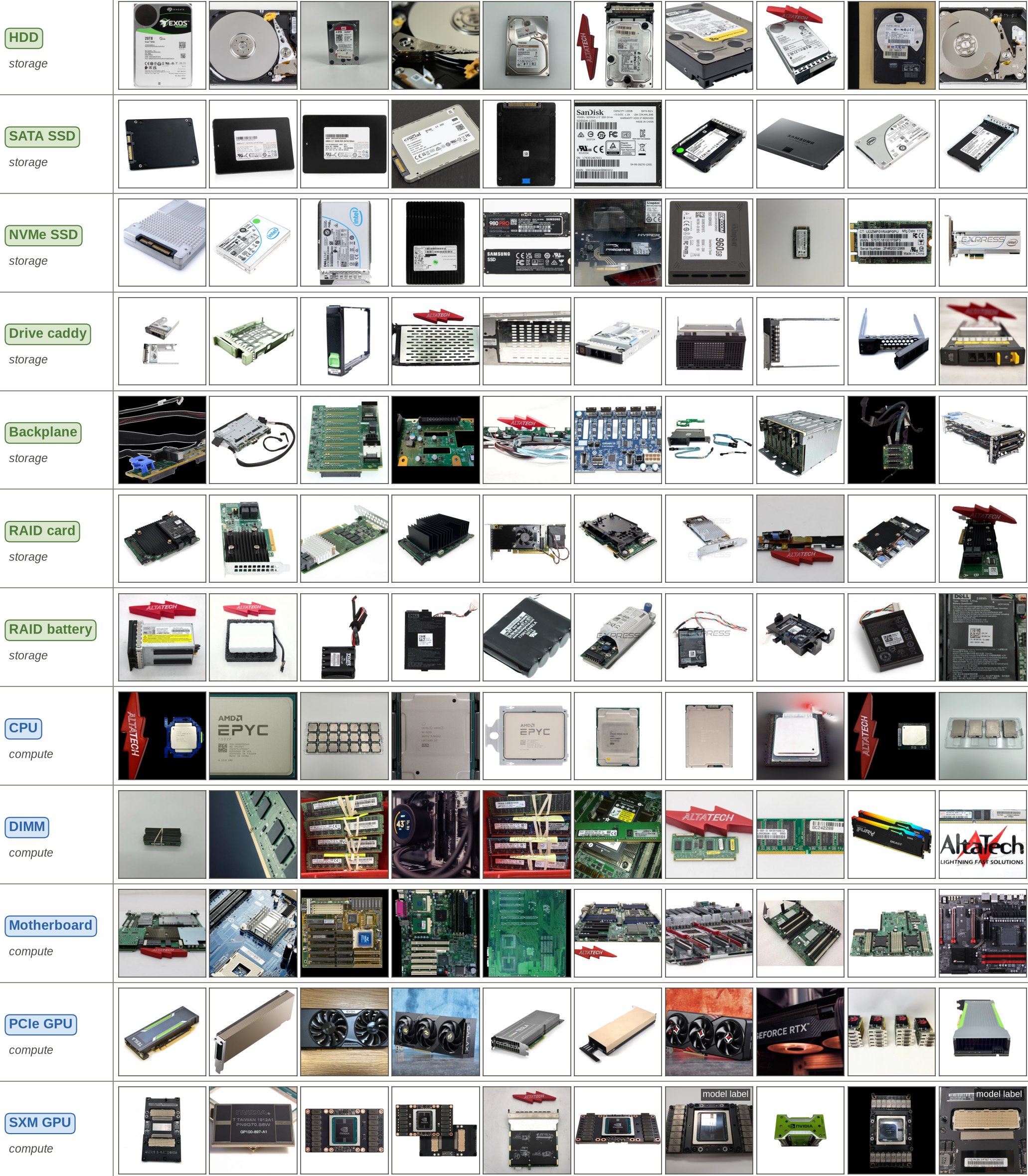}
  \caption{Ten images for each of the 24 classes, part 1: storage and compute. 238 are test images with a class given by human annotators: 193 on which two or more annotators agree and 45 with one confident answer that the model label matches. Images whose human class does not fit the class definition are not shown. The test set holds too few such SXM GPU modules, so two tiles of that row are training images with their model label, marked \emph{model label}. Images are center crops, at most three per source in a row where the sources allow it and one per near-duplicate group. Rows mix catalog shots, listing photographs with reseller props and overlays, and archive photographs.}
  \label{fig:gallery_a}
\end{figure*}
\begin{figure*}[tp]
  \centering
  \includegraphics[width=\linewidth]{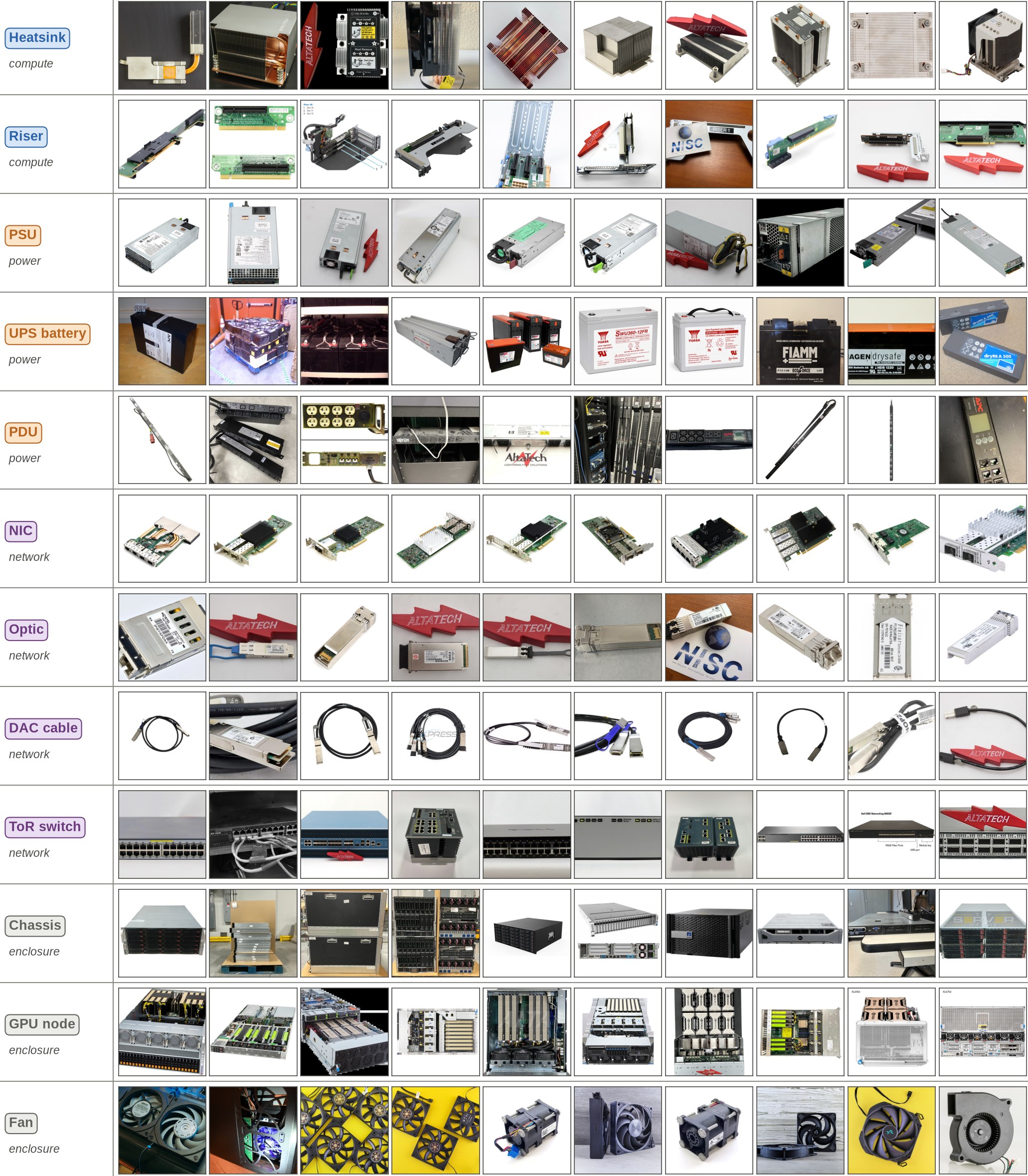}
  \caption{Ten test images for each of the other 12 classes, part 2: compute, power, network and enclosure. Selection as in \cref{fig:gallery_a}.}
  \label{fig:gallery_b}
\end{figure*}
\Cref{fig:gallery_a,fig:gallery_b} show ten images for each of the 24 classes, all but two of them test images with the class human annotators gave them.
Within a row the images come from several sources and from different near-duplicate groups, so the gallery shows how much one class varies between catalog shots, resale listings and archive photographs, which is the variation the classifiers of \cref{sec:cls} have to absorb.

\section{Datasheet}
\label{sec:supp_datasheet}
Following~\cite{gebru2021datasheets}.
\paragraph{Motivation.} Recognition of data-center hardware leaving service by class, condition, granularity and printed text, for sorting, reuse and recycling research (\cref{sec:intro}).
\paragraph{Composition.} 762,129 unique images in 389,169 near-duplicate groups. Splits: train 305,587, validation 12,652, test \ntestdrawn (100 controls, \ntestsent sent to annotators, \ntestscored with an adjudicated class), near-duplicates of test images withheld from every split 10,955, non-hardware photographs, a pool from which 100,000 are sampled as the \emph{none} class in training, 370,665, renders and graphics used only for the relevance filter 48,866, without a usable class 3,404 (\cref{tab:funnel}).
\paragraph{Collection.} Nine public datasets and collected images from public archives, manufacturer pages, refurbisher catalogs, surplus auctions and discovered sources, with per-image provenance (\cref{sec:collection}).
\paragraph{Preprocessing and labels.} Content-hash deduplication, perceptual-hash grouping, relevance filtering, pseudo-labels for train and validation, human labels for test (\cref{sec:curation,sec:labels}).
\paragraph{Uses, distribution, maintenance.} The dataset is intended for research on sorting, reuse and recycling of data-center hardware: classification, detection, condition and text recognition, and the study of label noise and leakage. Labels, boxes, text readings and metadata are released under CC BY-NC 4.0. An image is distributed only where its own license permits redistribution, and then under that license. For the other images the labels are released without the pixels, or the image is held back (\cref{fig:pipeline}). Access requires accepting terms of use that restrict the data to research and prohibit using it to locate equipment for theft or to recover data from storage devices. The dataset will be released at publication in a public repository with a persistent identifier. Every release is numbered and has a frozen manifest and a change log, and a contact address published with the release receives removal requests, which are logged and applied (\cref{sec:supp_ethics}).
Splits exist only in versioned release snapshots, and no synthetic image will ever enter validation or test.

\section{Why data-center hardware: life cycle and projected e-waste}
\label{sec:supp_motivation}
\begin{figure*}[tp]
  \centering
  \includegraphics[width=\linewidth]{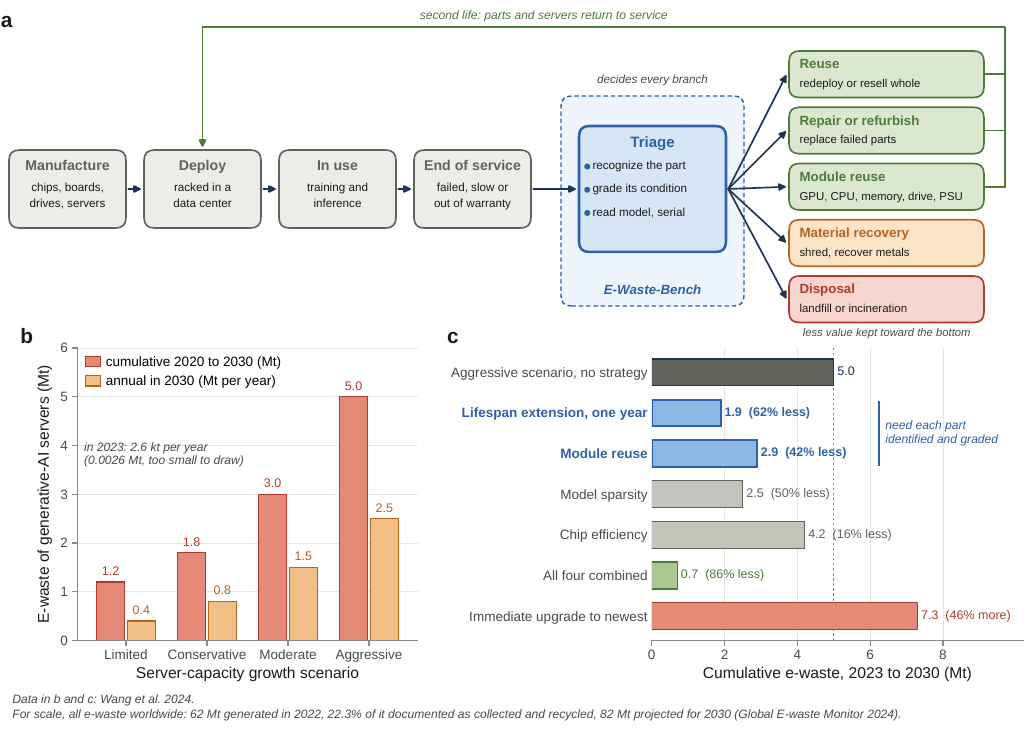}
  \caption{Why hardware leaving service must be recognized. (a) The life cycle of data-center hardware. Every route after end of service, from redeployment to disposal, starts with a triage step that recognizes the part, grades its condition and reads its identifier. That step is what \ewn targets. (b) Projected e-waste of generative-AI servers under four scenarios of server-capacity growth. (c) The cumulative e-waste that remains in the aggressive scenario under each circular strategy. The two highlighted strategies, lifespan extension and module reuse, depend on knowing what each part is and what state it is in. The schematic is ours. The numbers in (b) and (c) are from Wang et al.~\cite{wang2024genai}, and the global totals from the Global E-waste Monitor~\cite{balde2024gem}.}
  \label{fig:supp_lifecycle}
\end{figure*}
\Cref{fig:supp_lifecycle} places the dataset in the life cycle of the hardware and gives the scale of the problem it serves.
Global electronic waste is growing five times faster than documented recycling: the world generated 62 million tonnes of e-waste in 2022, only 22.3\% was documented as collected and recycled, an estimated US\$62 billion of recoverable resources were lost, and the total is projected to reach 82 million tonnes by 2030~\cite{balde2024gem}.
Data centers are a fast-growing part of it, since drives wear out, memory modules, power supplies and fans fail, optics and network cards are swapped on upgrades, and accelerators are replaced at every refresh, long before a server is retired.
In the United States their electricity use is projected to grow from 176~TWh in 2023 to between 325 and 580~TWh by 2028~\cite{shehabi2024datacenter}, and the servers behind generative AI alone could add 1.2 to 5.0 million tonnes of e-waste between 2020 and 2030, of which reuse, life extension and recycling could avoid 16 to 86\%~\cite{wang2024genai}.
Of the strategies studied by Wang et al.~\cite{wang2024genai}, extending a server's life by one year avoids 62\% of the projected e-waste and reusing its modules 42\%, and both begin with knowing which part is on the bench and what state it is in.

\begin{figure*}[tp]
  \centering
  \handfig[0.9\linewidth]{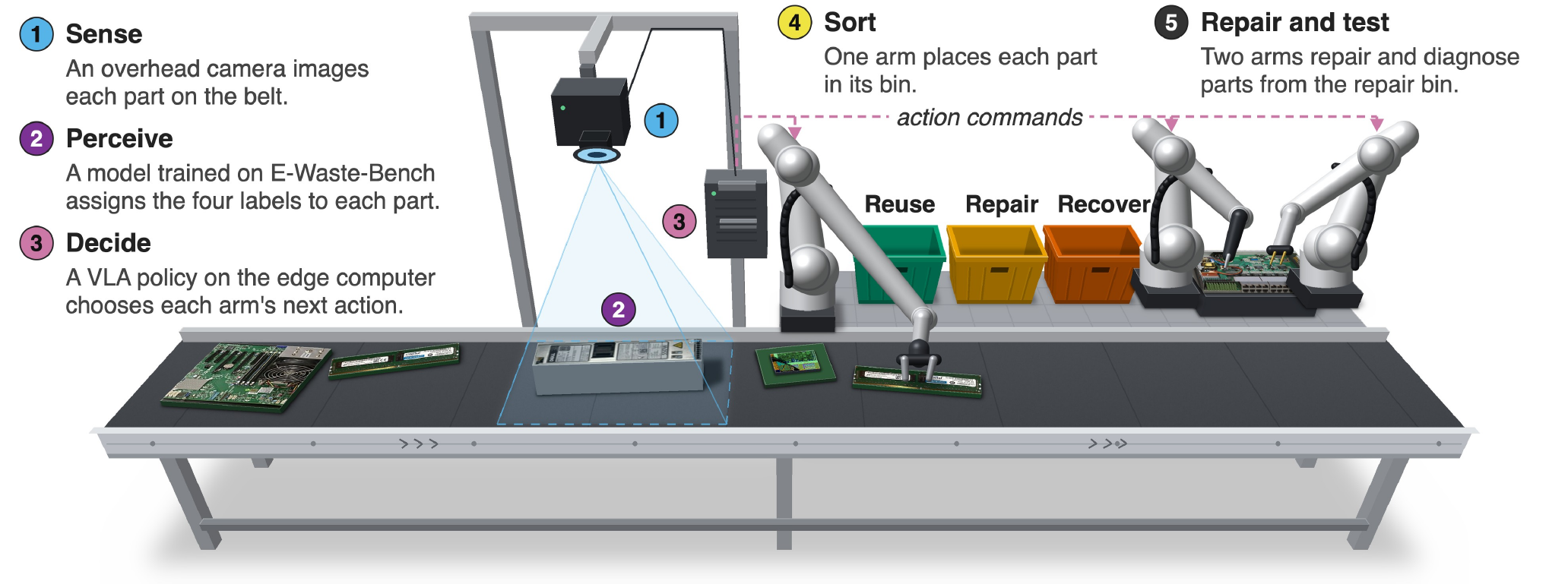}
  \caption{Envisioned use of \ewn: an automated triage cell for retired data-center hardware. The cell has not been built and is not evaluated in this paper, which supplies data and baseline models for step 2 only. Parts arrive on a conveyor in a mixed stream. (1) An overhead camera images each part as it passes under the gantry. (2) A perception model trained on \ewn assigns the four labels used throughout the paper: class, granularity, condition and printed text. (3) A vision-language-action (VLA) policy on an edge computer at the line acts on these labels and chooses the next action of each arm. The dashed line marks the action commands it sends to the arms. (4) One sorting arm places each part in a reuse, repair or material-recovery bin. Here it lifts an intact memory module for reuse, and a fractured GPU die waits on the belt for recovery. (5) Two arms on a test fixture repair and diagnose the parts from the repair bin, here a network switch with failed power-supply capacitors.}
  \label{fig:supp_usecase}
\end{figure*}

\Cref{fig:supp_usecase} sketches an envisioned use of the labels, a triage cell that sends each part to reuse, repair or recovery, and of its five steps this paper supplies data and baseline models for the second, perception.

\section{Images per class}
\label{sec:supp_classes}
\begin{table}[t]
\centering
\small
\setlength{\tabcolsep}{2.5pt}
\caption{Labeled hardware images per class, by split and by origin. Splits use the model-made class. Collected: the images E-Waste-Bench collected. Public: the nine public collections. Gain: collected images as a multiple of the images of the nine public collections combined.}
\label{tab:classes}
\begin{adjustbox}{max width=\linewidth}
\begin{tabular}{lrrrrrr}
\toprule
class & train & val & test & collected & public & gain \\
\midrule
HDD & 26,744 & 1,146 & 424 & 26,894 & 1,420 & 19$\times$ \\
SATA SSD & 11,154 & 560 & 425 & 11,418 & 721 & 16$\times$ \\
NVMe SSD & 4,176 & 113 & 408 & 3,639 & 1,058 & 3.4$\times$ \\
DIMM & 29,710 & 829 & 448 & 30,066 & 921 & 33$\times$ \\
PSU & 21,982 & 1,039 & 396 & 23,276 & 141 & 165$\times$ \\
Fan & 11,566 & 538 & 405 & 11,746 & 763 & 15$\times$ \\
PCIe GPU & 11,114 & 459 & 430 & 11,720 & 283 & 41$\times$ \\
SXM GPU & 604 & 34 & 156 & 793 & 1 & 793$\times$ \\
CPU & 11,710 & 455 & 494 & 11,989 & 670 & 18$\times$ \\
Motherboard & 11,550 & 480 & 393 & 11,183 & 1,240 & 9.0$\times$ \\
NIC & 24,064 & 987 & 421 & 25,420 & 52 & 489$\times$ \\
Optic & 6,748 & 204 & 445 & 7,175 & 222 & 32$\times$ \\
DAC cable & 10,020 & 440 & 369 & 10,755 & 74 & 145$\times$ \\
RAID card & 22,976 & 990 & 419 & 24,070 & 315 & 76$\times$ \\
RAID battery & 4,639 & 179 & 430 & 4,990 & 258 & 19$\times$ \\
UPS battery & 1,994 & 56 & 431 & 2,469 & 12 & 206$\times$ \\
Drive caddy & 17,040 & 790 & 348 & 18,090 & 88 & 206$\times$ \\
Backplane & 3,210 & 123 & 417 & 3,739 & 11 & 340$\times$ \\
Riser & 8,576 & 351 & 421 & 9,050 & 298 & 30$\times$ \\
Heatsink & 9,498 & 476 & 455 & 9,982 & 447 & 22$\times$ \\
Chassis & 38,582 & 1,451 & 464 & 39,972 & 525 & 76$\times$ \\
GPU node & 109 & 5 & 75 & 165 & 24 & 6.9$\times$ \\
PDU & 1,825 & 73 & 290 & 2,186 & 2 & 1,093$\times$ \\
ToR switch & 15,996 & 579 & 440 & 16,764 & 251 & 67$\times$ \\
\midrule
all 24 classes & 305,587 & 12,357 & 9,404 & 317,551 & 9,797 & 32$\times$ \\
\bottomrule
\end{tabular}
\end{adjustbox}
\end{table}

\begin{figure*}[tp]
  \centering
  \includegraphics[width=\linewidth]{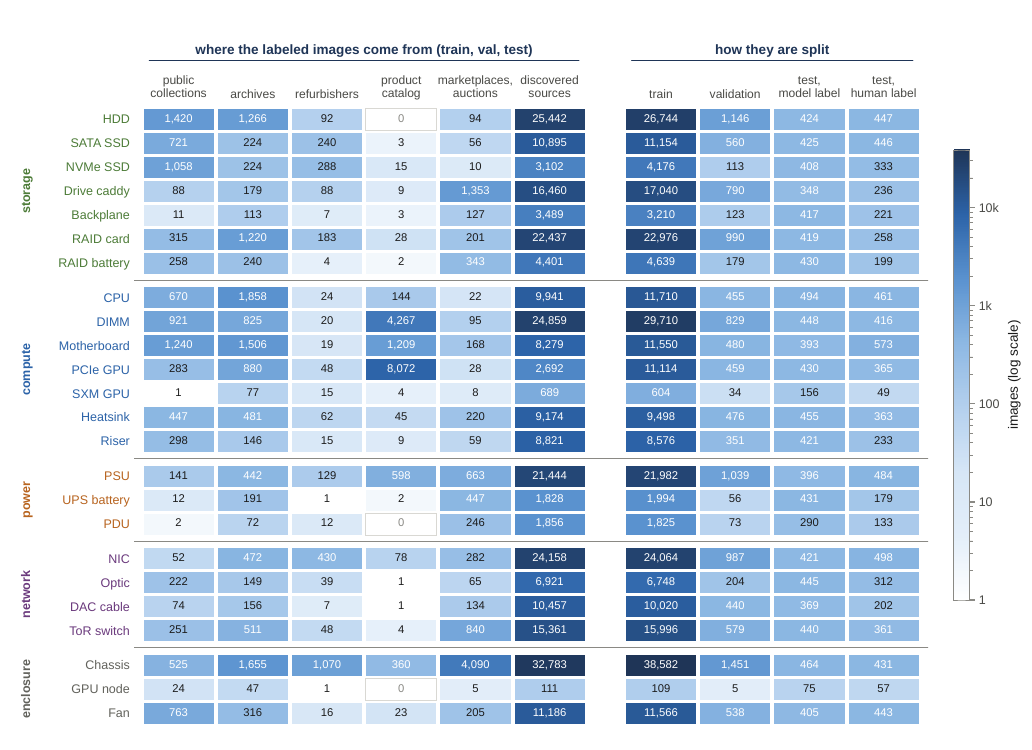}
  \caption{Where the images of each class come from and how they are split. Rows are the 24 classes by function group. Left block: labeled hardware images of the training, validation and test splits by kind of source. Right block: the same images by split, with the test set counted twice, by the class the models gave and by the class human annotators gave. Shading is logarithmic and each cell prints its count. The public collections column is the gap the collection closes: they hold one SXM GPU module, 2 PDUs and 11 backplanes.}
  \label{fig:supp_heatmap}
\end{figure*}
\Cref{tab:classes} gives the labeled hardware images of each class by split and by origin, and \cref{fig:supp_heatmap} breaks the same counts down by kind of source and by split.
The public collections column of the heatmap shows where the collection adds most, in the classes those collections barely hold, such as SXM GPU modules, PDUs and backplanes.
The split columns of \cref{tab:classes} count images by their model-made class, so its test column (9,404) is the part of the \ntestdrawn drawn test images that the pseudo-labels place in one of the 24 classes, not a count of human labels, and its validation column (12,357) is likewise the part of the 12,652 validation images with a pseudo-label class.
The classes are field-replaceable units that fleet studies document as failing or replaced in service, including drives~\cite{backblaze2026drivestats,miller2023hdd,maneas2020ssd}, memory~\cite{beigi2023ddr4}, GPUs and GPU nodes~\cite{grattafiori2024llama3,kokolis2025reliability}, CPUs~\cite{wang2023sdc}, network switches and transceivers~\cite{meza2018network,zhuo2017corropt}, and RAID cards, power supplies, fans and backplanes~\cite{vishwanath2010characterizing,wang2017fouryears}.
They also cover parts handled in the same swaps (drive caddies, risers, heatsinks, RAID batteries) and high-value parts recovered at decommissioning~\cite{wang2024greensku}.

\section{Collection and source policy}
\label{sec:supp_collection}
Sources fall into tiers by kind: institutional and Creative Commons archives, manufacturers, a product catalog, refurbishers, marketplaces and auctions, technical forums, and hosts discovered automatically from pages already read.
Collection reads only content that is publicly available or offered through an official API, and passes no login, paywall or other access control.
Every request goes through one fetch path that enforces a per-host limit of 20 requests per minute, checks that every redirect resolves to a public address, allows only http and https, identifies the crawler honestly and sends an honest referrer.
A per-source policy records the permitted collection methods (official API, product feed, plain HTTP), the access basis and the retention scope, and a source that denies access is paused, never retried under another identity.
Store product feeds and official APIs are preferred where a source offers them.
The labeled images come from 143 web sources: 4 public archives (13,817 images), 4 refurbisher catalogs (2,859), a product catalog (14,893), an auction site (9,769), a manufacturer and a forum (under 20 images together), and 131 sources discovered by the pipeline (277,085), beside the public collections (9,797).

Robots.txt was followed for institutional, manufacturer, refurbisher and forum hosts.
For one product catalog, for marketplaces and auction sites, and for automatically discovered hosts it was not applied, a decision approved by the project lead, the faculty advisor and the university, and these sources were read under the same rate limit and identification.
A full licensing, IRB and legal review of the collected images follows collection and is completed before release: an image is distributed only where its license permits, the labels of the other images are released without their pixels or the image is held back, and a logged process honors removal requests and blocks the source (\cref{sec:supp_ethics}).
Every manifest row records its source, source URL, the query or listing that led to it, the collection time and the licensing fields.

\section{Annotation protocol}
\label{sec:supp_annotation}
\begin{figure*}[tp]
  \centering
  \begin{minipage}[b]{0.652\linewidth}
    \centering
    \includegraphics[width=\linewidth]{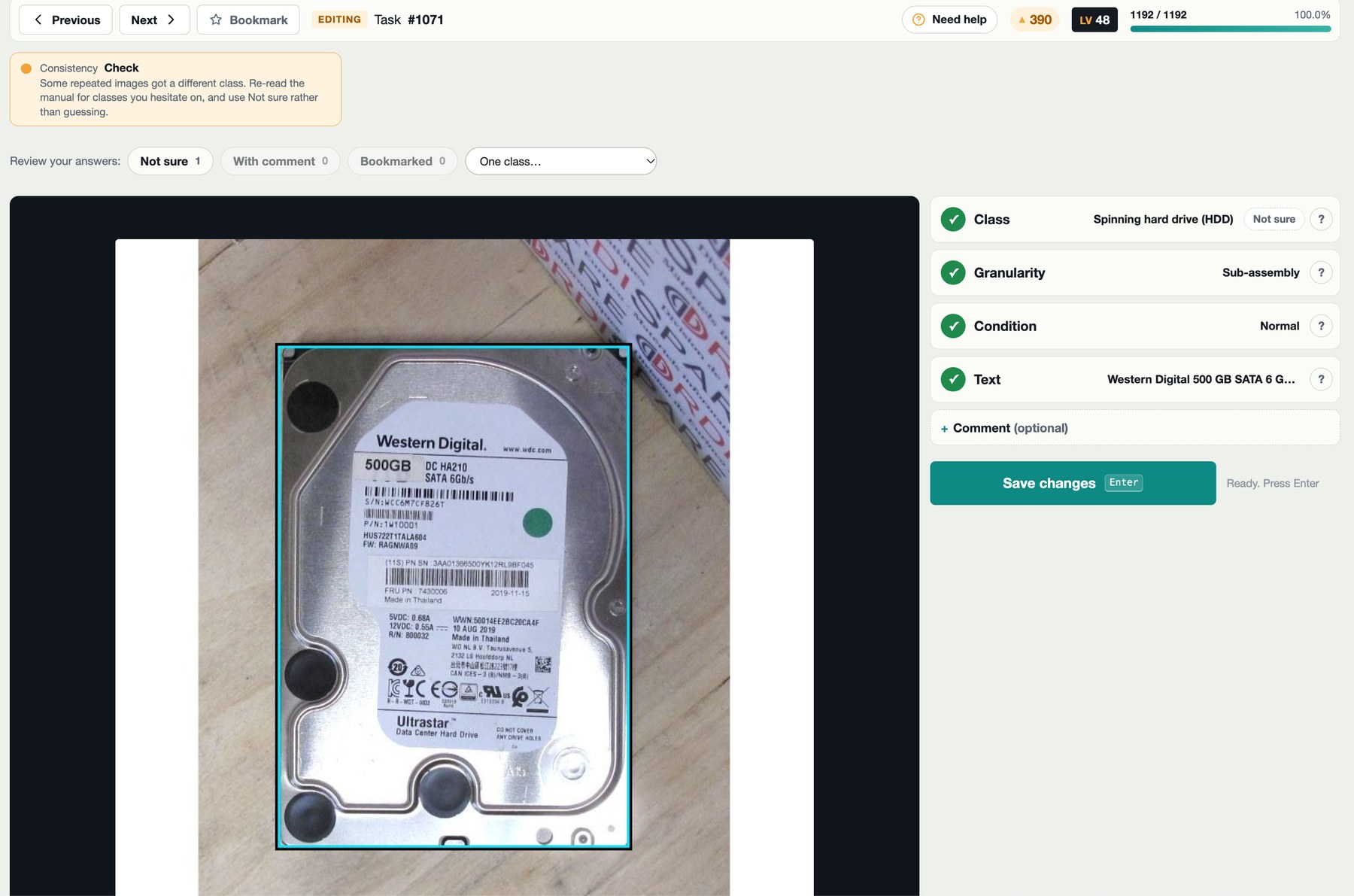}\\[2pt]
    {\small (a) one task, the whole screen}
  \end{minipage}\hfill
  \begin{minipage}[b]{0.326\linewidth}
    \centering
    \includegraphics[width=\linewidth]{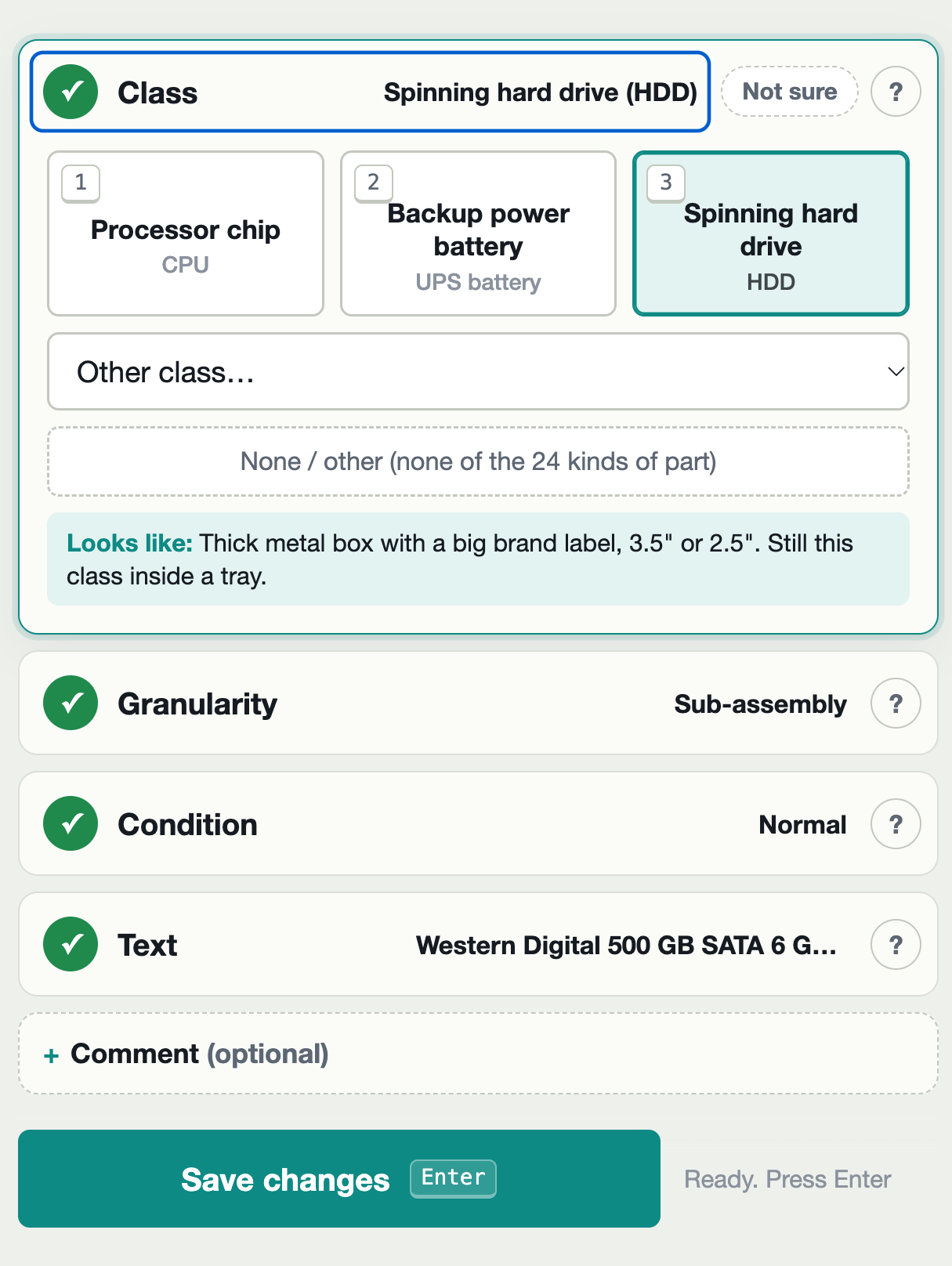}\\[2pt]
    {\small (b) the class step opened}
  \end{minipage}\\[8pt]
  \includegraphics[width=\linewidth]{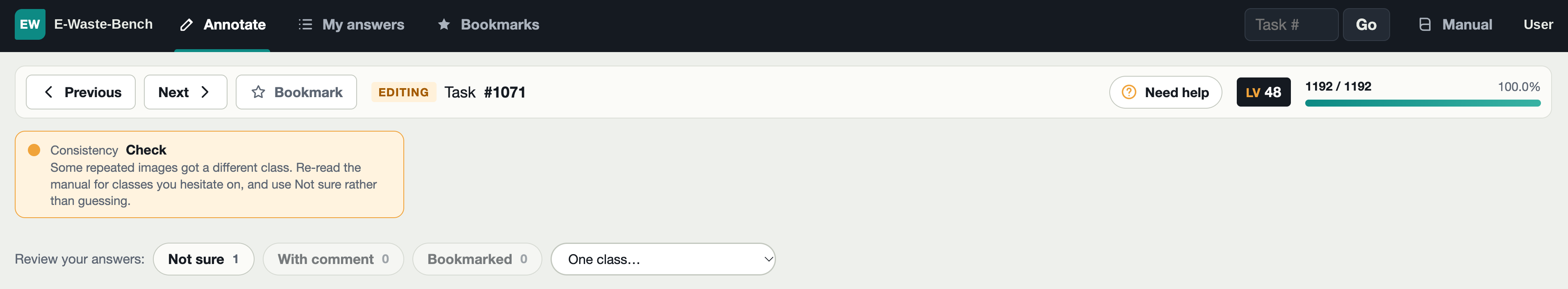}\\[2pt]
  {\small (c) the top of the page}
  \caption{The annotation tool. (a) One task, here reopened for editing with the four answers collapsed. The image is shown with its main object outlined, the only object the answers are about. (b) The class step opened: three suggested classes in random order on keys 1 to 3 (here the third is chosen), a list of all other classes, the \emph{None / other} button, a short description of the chosen class and the \emph{Not sure} flag. Below it the other three steps, \emph{Granularity}, \emph{Condition} and \emph{Text}, answered and collapsed, each with a \emph{?} button that opens the matching section of the manual, an optional \emph{Comment}, and the submit button (also the Enter key), which turns on once all four steps are answered. (c) The navigation bar (\emph{Annotate}, \emph{My answers}, \emph{Bookmarks}, a task-number box and the \emph{Manual}), the task bar (\emph{Previous} and \emph{Next}, \emph{Bookmark}, the task number, \emph{Need help}, which sends the organizers a question with the details of the current image, and the progress bar), the \emph{Consistency Check} notice, shown when repeated images received a different class, and the \emph{Review your answers} row, which filters the annotator's own tasks by \emph{Not sure}, \emph{With comment}, \emph{Bookmarked} or one class. Panels (b) and (c) are screenshots of the same task at a higher resolution, and the account name in (c) is replaced by \emph{User}.}
  \label{fig:supp_site}
\end{figure*}
Annotators work on a web tool in batches of 50 images (\cref{fig:supp_site}).
Each task shows one image with its main object outlined, and the answers are about that object only: if the outline frames a drive lying on a motherboard, the answer is about the drive.
Every task gets four answers in a fixed order.
(1) \emph{Class}. The tool offers three model suggestions as buttons, in random order, beside the list of all 24 classes and a \emph{none} button. The manual tells annotators that the suggestions are often wrong on the hard images, and to look and decide first and use a button only if it matches. \emph{Not sure} marks the class as uncertain and asks for a second candidate.
(2) \emph{Granularity}: device, sub-assembly or component, or none when the object is not hardware at all.
(3) \emph{Condition}: every value that applies. \emph{Normal} stands alone, so selecting it clears the others and selecting any other clears it.
(4) \emph{Text}: the text printed on the object, or a tick on \emph{No readable text}. Where the OCR reader and the vision-language reader found text, each reading is shown with a button that copies it into the box, to be used only if it is exactly right and corrected otherwise.
A typed text was not marked when it came from a copied line, and 931 of the 3,344 typed texts (27.8\%) are character for character a line the tool showed.
The found rates of the two readers in main-paper \cref{sec:text} therefore include lines annotators kept.
Gemma~3, whose readings no annotator saw, contains the identifier on 37.6\% of the images it read, close to the OCR reader (38.5\%) and Qwen3.5 (41.7\%) on the same images (\cref{sec:supp_tables}), so the copying moves these rates by a few points at most.
The submit button turns on once all four steps are answered, and an optional \emph{Comment} can be added below them.
Keys 1 to 3 pick a suggestion, U marks \emph{Not sure}, B sets a \emph{Bookmark} and Enter submits.
An earlier answer can be reopened and edited through \emph{Previous}, the review row or its task number, and the tool keeps the first answer of every edited task.
Once an image reaches an annotator for the second time, the earlier answer to it disappears from the annotator's history and can no longer be reopened, so a repeat cannot be compared with it or made to match. Repeats are spread evenly over each account's tasks.
The manual tells annotators that some images return, to answer each one as if it were new, and that their consistency with themselves is part of quality control.

\begin{figure*}[tp]
  \centering
  \includegraphics[width=\linewidth]{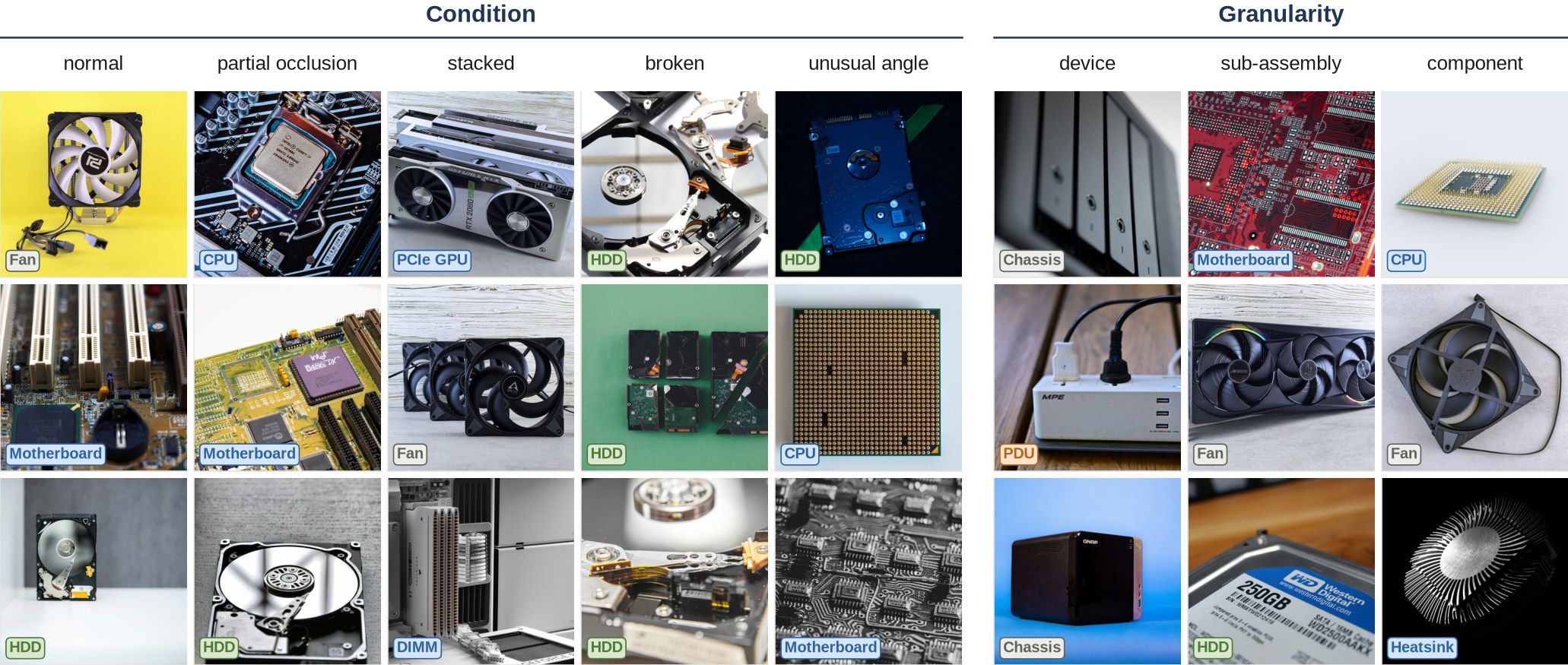}
  \caption{What the condition and granularity labels look like. One column per value, three test images each, with the class human annotators gave. Every image carries the column's value in every answer the annotators gave for it, and the definitions annotators worked from are in \cref{tab:supp_defs}. Only images whose license allows reproduction are shown. The sixth condition value, \emph{deformed}, has no column because no test image that may be reproduced carries it in every answer, and its definition is in \cref{tab:supp_defs}.}
  \label{fig:label_axes}
\end{figure*}
\begin{figure}[tp]
  \centering
  \includegraphics[width=\linewidth,trim=144.5 8.1 144.5 8.1,clip]{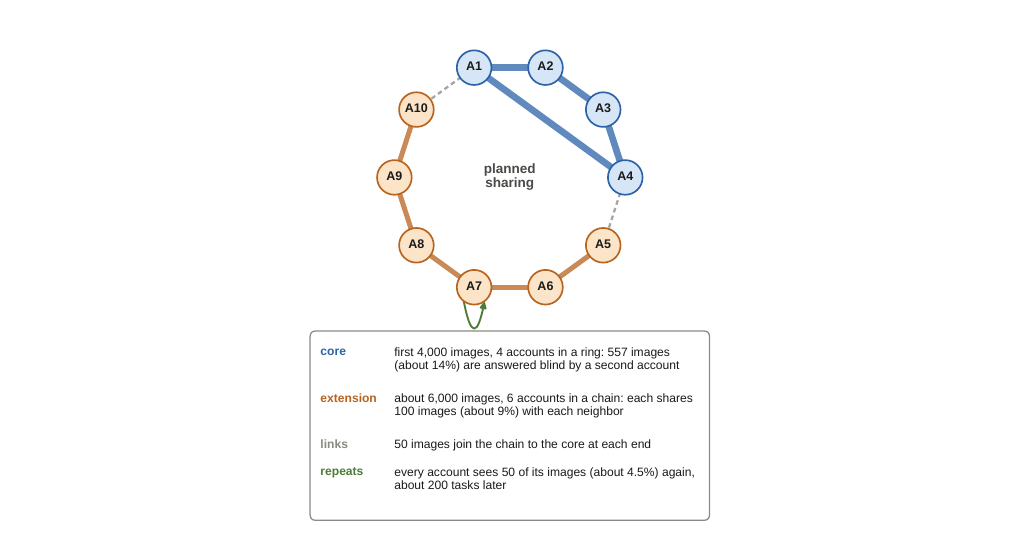}
  \caption{How part of the test set was planned to be answered twice. The first 4,000 images go to four accounts, and about 14\% of them (557 images) are answered blind by a second account. The extension (about 6,000 images) goes to six accounts in a chain linked to the core at both ends. Each of these shares 100 images (about 9\%) with each neighbor, and every account sees 50 of its images (about 4.5\%) again about 200 tasks later. What was measured is main-paper \cref{fig:annotation_overlap}.}
  \label{fig:annotation_design}
\end{figure}

\begin{table*}[t]
\centering
\small
\caption{The definitions annotators worked from, condensed from the annotation manual. Condition values can be combined, except \emph{normal}.}
\label{tab:supp_defs}
\begin{tabular}{@{}llp{0.40\linewidth}p{0.30\linewidth}@{}}
\toprule
axis & value & definition & boundary cases \\
\midrule
condition & normal & The whole object is visible, undamaged and shown from an ordinary view. Chosen only when no other value applies. & Dust, scratches, stickers and price labels do not prevent it. \\
 & partial occlusion & About a fifth or more of the object is hidden by another object, a hand, packaging or a cable, or cut off by the edge of the photo. & A drive half inside its bay counts. Labels on the object and a small cut corner do not. \\
 & stacked & The object is part of a pile, stack, row or bulk lot of similar items that touch or overlap. & A tray of DIMMs or servers in a rack count. Two different items side by side do not. \\
 & broken & Material is cracked, snapped, missing, burnt or destroyed. & Drives destroyed for data security count. An empty slot or a server without drives does not. \\
 & deformed & The shape is bent, warped, dented, crushed, twisted or swollen, even if nothing has broken off. & Bent socket pins, bent brackets and swollen batteries count. \\
 & unusual angle & The viewpoint hides the usual identifying side: extreme oblique views, from below or behind, upside down, or an extreme close-up. & The test: would a technician need a second look to recognize it? \\
\midrule
granularity & device & A complete, standalone unit with its own enclosure. & An open or stripped server is still a device. \\
 & sub-assembly & A replaceable unit built from several parts and removed as one piece, such as a drive, a power supply or a memory module. & \\
 & component & A single part, usually not taken apart further in the field, such as a CPU, a heatsink or a transceiver. & A bare fan without its carrier is a component. \\
\bottomrule
\end{tabular}
\end{table*}
\Cref{fig:label_axes} shows examples of every condition and granularity value, and \cref{tab:supp_defs} gives the definitions.
Most classes have one typical granularity (\cref{tab:supp_defs}), and annotators depart from it only when the outline shows something more or less complete than usual.
For text, annotators copy brand names, model and part numbers, capacity, speed and power values and type markings exactly as printed. They leave out serial numbers, asset tags, barcode digits, MAC addresses and other unique identifiers, watermarks and seller overlays, text on other objects, and characters they cannot read with confidence.
\emph{None} is for an object that is not one of the 24 kinds of part or cannot be recognized. The guide lists whole consumer devices (laptops, phones, monitors, printers), complete UPS units, routers, firewalls and access points, internal SAS, SATA and ribbon cables, patch cables and power cords, consumer batteries and chargers, laptop and phone logic boards, memory cards, bare chips and loose screws or brackets, drawings, diagrams and screenshots, and an outline that frames no clear object or several unrelated items.
The full manual given to the annotators, with its example images for every class and its lists of look-alikes and difficult cases, will be released with the dataset.
The overlap design is shown in \cref{fig:annotation_design} and the measured overlap in main-paper \cref{fig:annotation_overlap}: 557 of the first 4,000 images (about 14\%) are answered blind by a second account, the six accounts of the extension form a chain in which each shares 100 images with each neighbor, and every account sees 50 of its images again (25 in each of the two accounts that came from a split queue). In the order an account answered, the repeat comes at least 193 tasks after the first view in eight accounts, as early as 77 tasks in one, and after about 25 tasks in the two accounts that came from a split queue (A6 and A11).
Two accounts were replaced and one was split during annotation (A4, A8, and A6 into A6 and A11), so main-paper \cref{fig:annotation_overlap} has no edge from A8 to A9 and a thin one from A6 to A7.
Ten annotators were recruited. Two withdrew, and their queues went to two of the remaining annotators with the place of each queue in the design unchanged, so its shared images, repeats and neighbors are the ones planned. With one queue that was split, three of the eight annotators therefore hold two accounts each (11 accounts in all), and answers given under two accounts of one person are never counted as two annotators.
An answer can be flagged as unsure, and a confident answer then decides the image's class. Otherwise the majority decides, and a tie between two answers goes to a blind third vote by a co-author who saw neither earlier answer.
Agreement per axis is in \cref{tab:agreement}: it compares first answers for repeats by the same annotator, so an answer later edited to match does not count.

\begin{table}[t]
\centering
\small
\setlength{\tabcolsep}{3pt}
\caption{Agreement on the test images that were answered twice. Class and granularity: percent agreement and Cohen's $\kappa$. Condition: Krippendorff's $\alpha$ per value. Same annotator: a repeat of the image, typically about 200 tasks later (spacing per account in \cref{sec:supp_annotation}), first answers for class and granularity, final answers for condition (498 repeats with a condition on both).}
\label{tab:agreement}
\begin{adjustbox}{max width=\linewidth}
\begin{tabular}{lrr}
\toprule
 & two annotators & same annotator \\
\midrule
class, agreement (\%) & 74.0 & 84.0 \\
class, Cohen's $\kappa$ & 0.72 & 0.83 \\
granularity, agreement (\%) & 73.8 & 78.8 \\
granularity, Cohen's $\kappa$ & 0.60 & 0.67 \\
\midrule
condition: normal, $\alpha$ & 0.23 & 0.52 \\
condition: partial occlusion, $\alpha$ & 0.17 & 0.56 \\
condition: stacked, $\alpha$ & 0.50 & 0.74 \\
condition: broken, $\alpha$ & 0.11 & 0.00 \\
condition: deformed, $\alpha$ & 0.18 & 0.00 \\
condition: unusual angle, $\alpha$ & 0.20 & 0.27 \\
\midrule
pairs of answers & 1,057 & 500 \\
\bottomrule
\end{tabular}
\end{adjustbox}
\end{table}

\begin{figure*}[tp]
  \centering
  \includegraphics[width=\linewidth]{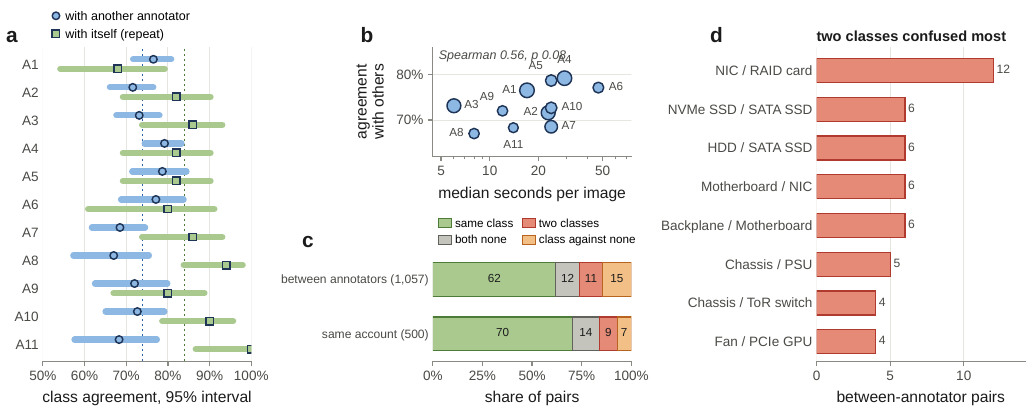}
  \caption{How well human annotators agree on the test set. (a) Class agreement of each annotator account with another annotator and with itself when an image returns, with 95\% Wilson intervals. Dashed lines are the pooled values. (b) Median seconds per image against agreement with others, one dot per account, sized by its number of pairs. (c) The kinds of answer pairs: 56\% of the disagreements between annotators are one of them answering \emph{none}. (d) The class pairs annotators confuse most often, which are parts that look alike. Counts in (d) are small and show which parts are confused, not how often.}
  \label{fig:annotation_agreement}
\end{figure*}
\begin{table*}[t]
\centering
\small
\caption{Answers and agreement per annotator account on the test set. Between: pairs with another person on the same image. Self: the same image shown again to the same account, at least 193 of its own tasks later in eight accounts, 77 in A3 and about 25 in A6 and A11, first answers compared. No image was answered by two accounts of one person (A4 and A10, A5 and A11, A8 and A9), so every between pair is two annotators. Time: median seconds from first view to first save, gaps over 10 minutes left out as breaks. Best agreement in bold among accounts with at least 20 pairs.}
\label{tab:annotators}
\begin{adjustbox}{max width=\linewidth}
\begin{tabular}{lrrrrrrrr}
\toprule
account & answers & not sure (\%) & none (\%) & between pairs & between agr. (\%) & self pairs & self agr. (\%) & seconds per image \\
\midrule
A1 & 1,236 & 1.1 & 21.9 & 328 & 76.5 & 50 & 68.0 & 17 \\
A2 & 1,192 & 0.1 & 13.6 & 285 & 71.6 & 50 & 82.0 & 23 \\
A3 & 1,143 & 24.0 & 22.6 & 279 & 73.1 & 50 & 86.0 & 6 \\
A4 & 1,186 & 9.3 & 30.3 & 322 & \textbf{79.2} & 50 & 82.0 & 29 \\
A5 & 1,100 & 0.4 & 23.9 & 150 & 78.7 & 50 & 82.0 & 24 \\
A6 & 575 & 3.8 & 22.6 & 118 & 77.1 & 25 & 80.0 & 47 \\
A7 & 1,150 & 0.0 & 23.5 & 200 & 68.5 & 50 & 86.0 & 24 \\
A8 & 1,050 & 0.1 & 5.5 & 100 & 67.0 & 50 & 94.0 & 8 \\
A9 & 1,150 & 0.4 & 12.1 & 100 & 72.0 & 50 & 80.0 & 12 \\
A10 & 1,100 & 5.7 & 31.0 & 150 & 72.7 & 50 & 90.0 & 24 \\
A11 & 575 & 0.0 & 26.3 & 82 & 68.3 & 25 & \textbf{100.0} & 14 \\
\midrule
all & 11,457 & 4.3 & 21.0 & 1,057 & 74.0 & 500 & 84.0 & 17 \\
\bottomrule
\end{tabular}
\end{adjustbox}
\end{table*}

\Cref{fig:annotation_agreement} and \cref{tab:annotators} report the agreement per account, and four observations follow from them.
First, ten of the eleven accounts agree more with themselves than with others, and over all accounts self-agreement is 84.0\% over 500 repeats against 74.0\% over 1,057 pairs.
Second, 56\% of the disagreements between annotators are one annotator answering \emph{none} while the other names a class, so most of the disagreement is about whether an image belongs in the taxonomy at all, and the share of \emph{none} answers runs from 5.5 to 31.0\% across accounts.
Third, the \emph{not sure} flag marks hard images: pairs with a flagged answer agree on 47.1\% (104 pairs), confident pairs on 76.9\% (953 pairs).
Fourth, agreement depends on who meets whom, from 64.3 to 82.4\% over the ten account pairs that share at least 20 images.
Slower accounts tend to agree more with others (Spearman 0.56 over 11 accounts, $p = 0.08$).
Leaving out the 1,713 images answered only by the two fastest accounts (A3 and A8) moves ConvNeXt-B top-1 from 65.0 to 64.8\%.
The median time per image is 17 seconds.
The \ntestsent images were given out as 11,457 tasks: every image once, 1,057 images a second time to another annotator (blind overlap) and 500 again to the same annotator. The allocation planned 1,157 overlap images, and 100 were dropped when one account took over the queue of another, because its owner had already answered them.
Every score uses the \ntestscored images with an adjudicated class, and the 139 images still tied are left out of the scores.
A co-author who saw neither answer cast a blind third vote on 214 of the 286 ties, which settled 147. %
In 67 the vote matched neither answer, and the 72 ties that arose from answers given after the third vote was organized had none.
Counting all 139 as errors would lower ConvNeXt-B top-1 from 65.0 to 64.1\%.
The test draw of main-paper \cref{sec:labels} takes up to 444 images per class of the pseudo-labels it was drawn on (\cref{tab:classes} counts the final pseudo-labels, which place up to 494 test images in a class), no source supplies more than a quarter of a class except in the two rarest classes, and the disputed images are drawn on purpose so that the test set is not limited to what the pseudo-labelers get right.

\section{Every source of labels against human annotators}
\label{sec:supp_labelers}
\begin{figure*}[tp]
  \centering
  \includegraphics[width=\linewidth]{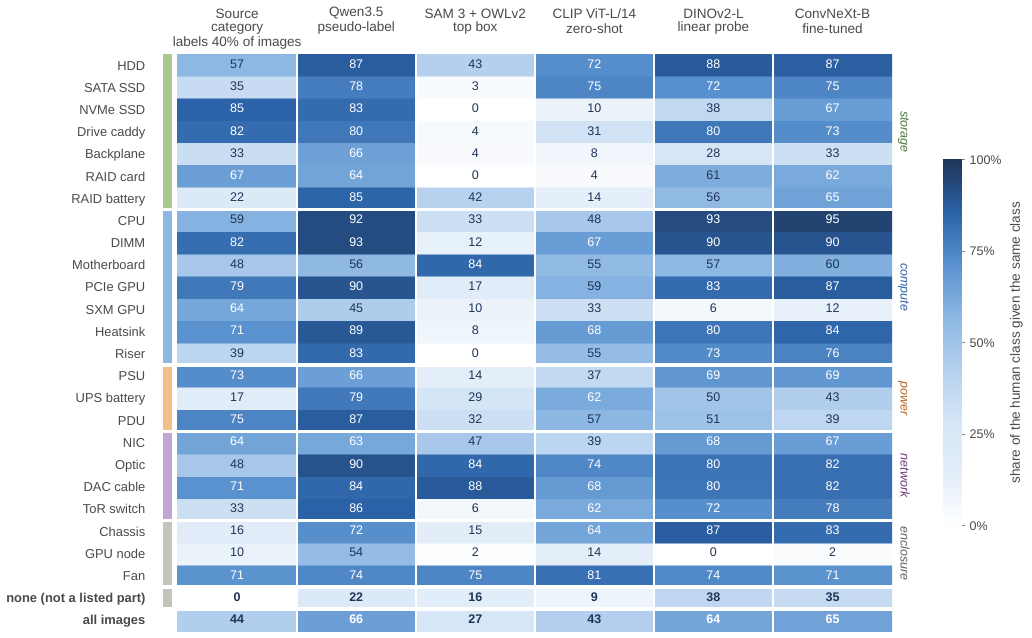}
  \caption{Every source of labels against the human annotation of the test set. A cell is the share of the test images of a human class (rows, grouped by function) to which a labeler (columns) gave the same class, in percent. The source category is what the web page or the search job said. It exists for 40\% of the images, so its cells cover those images only, and it can never say \emph{none}. The Qwen3.5 pseudo-label used for training agrees with the annotators on 66\% of all images but on only 22\% of the images they marked \emph{none}, and the classifiers trained on it inherit that weakness. The top-scoring box of the detectors is a poor image label (27\%), which is why the class of an image comes from the vision-language model and not from the boxes.}
  \label{fig:supp_labelers}
\end{figure*}
\begin{figure*}[tp]
  \centering
  \includegraphics[width=\linewidth]{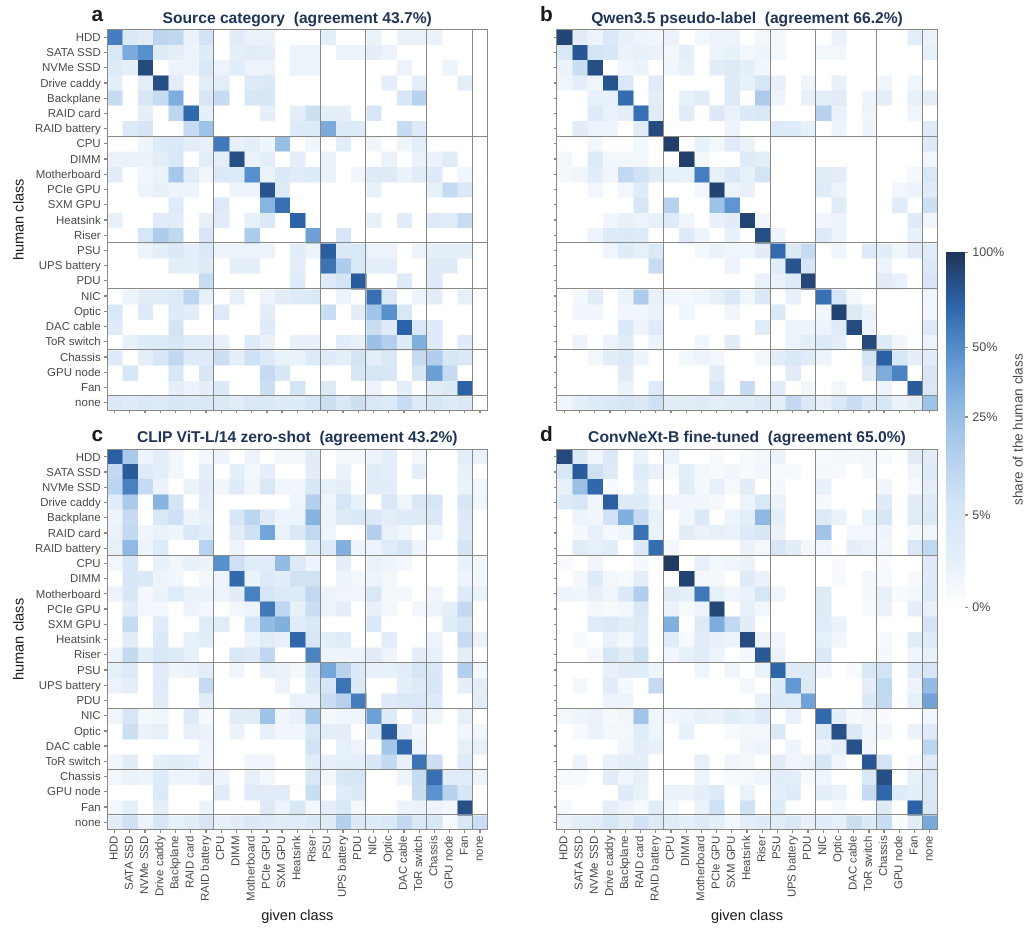}
  \caption{Confusion against the human class for four sources of labels: (a) the source category, (b) the Qwen3.5 pseudo-label, (c) CLIP ViT-L/14 zero-shot and (d) ConvNeXt-B trained on the pseudo-labels (three seeds pooled). Rows are human classes, columns the class given, each row sums to 100\%, and the color scale is compressed so that small shares off the diagonal stay visible. The bottom row shows where the images that human annotators marked \emph{none} go. The classifier repeats some confusions of the pseudo-labeler it was trained on (network card taken for RAID card) and adds its own (NVMe taken for SATA drive, backplane for riser).}
  \label{fig:supp_confusions}
\end{figure*}
\Cref{fig:supp_labelers,fig:supp_confusions} compare every source of labels we have with the human class of the \ntestscored scored test images.
The source category is right on 44\% of the images it labels, the pseudo-label on 66\% of all images, CLIP zero-shot on 43\%, a linear probe on DINOv2-L features on 64\% and the fine-tuned ConvNeXt-B on 65\%.
Among the 24 classes, the pseudo-label is weakest on SXM GPU modules (45\% of 49 images), GPU server nodes (54\% of 57) and motherboards (56\% of 573).
The \emph{Human annotators} row of main-paper \cref{tab:cls} is measured on the 1,057 images two annotators answered, which differ from the full test set, so it is not directly comparable with the models' top-1.
On the 782 \emph{agreed} images, where an annotator scores 100 by construction, Qwen3.5 reaches 80.4\% and Gemma 3 73.3\%.
That subset is small and holds the easier images, and annotators saw Qwen3.5 suggestions, so it bounds the label-noise effect only roughly.

\section{Scores on model-made labels and on human labels}
\label{sec:supp_silvergold}
\begin{figure*}[tp]
  \centering
  \includegraphics[width=\linewidth]{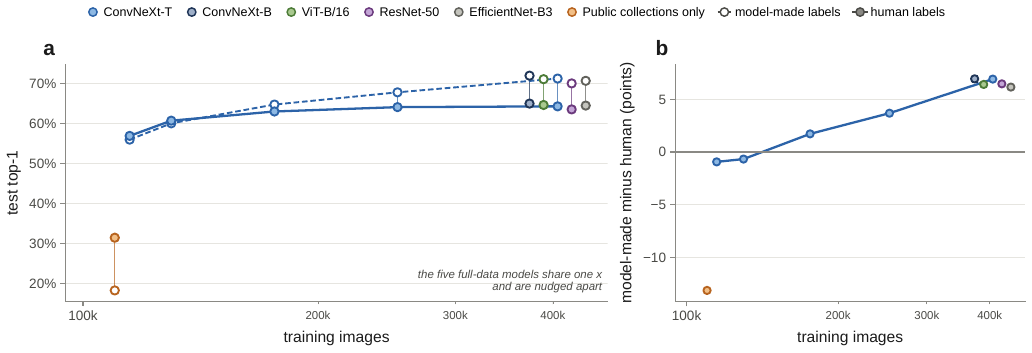}
  \caption{Scores on model-made labels overstate accuracy against human annotators, and more so as training data grows. (a) Test top-1 of every trained classifier on model-made labels (hollow) and on human labels (filled), on the same \ntestscored scored test images, against the number of training images. The ConvNeXt-T runs on 5, 10, 25, 50 and 100\% of the training groups are joined, and error bars are the standard deviation over three seeds where three were run. (b) The difference between the two scores. It grows from about zero at 5\% of the data to 7 points at full data: more pseudo-labeled data raises the score on model-made labels much faster than the score on human labels. The model trained on the public collections alone scores higher on human labels, as do the 5\% and 10\% runs by under one point.}
  \label{fig:supp_silvergold}
\end{figure*}
\Cref{fig:supp_silvergold} is the reason this paper reports no benchmark result on model-made labels and uses them only to measure how far they overstate accuracy.
At full data the five fine-tuned classifiers score 71.0\% on model-made labels and 64.4\% on human labels on average, a gap of 6.6 points.
The best and the worst model are the same under both label sets. The models between them change order, but they differ by at most 0.3 points on human labels, which is within the spread over seeds, so we do not read a reordering into it.
The runs below full data have one seed each.

\section{More on classification errors}
\label{sec:supp_cls}
\begin{figure*}[tp]
  \centering
  \includegraphics[width=\linewidth]{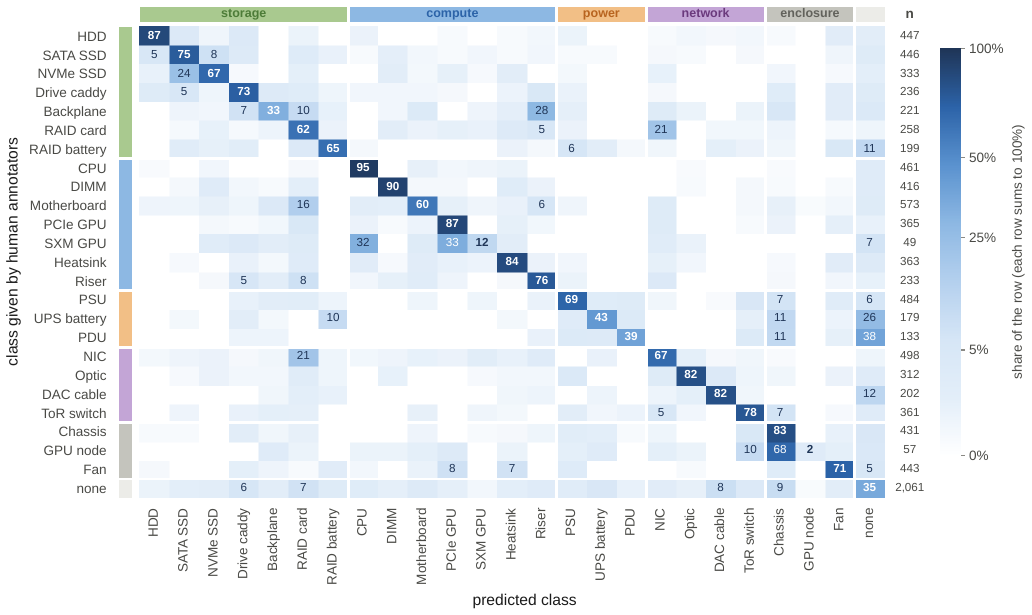}
  \caption{Confusion matrix of the best trained classifier, ConvNeXt-B, on the \ntestscored human-labeled test images, pooled over three seeds. Rows are the classes given by human annotators, columns the classes the model predicted, both grouped by function (the colored strips). Each row is normalized to percent, so it sums to 100: the diagonal cell is the accuracy on that class, and the cells beside it show where the rest of its images went. Off-diagonal cells under 5\% are not printed, and the column $n$ gives the number of test images in each row. The \emph{none} row holds the images the annotators marked as showing no data-center part, and the \emph{none} column the images the model rejected as showing none. Most confusions are between parts that look alike, such as NVMe and SATA drives or RAID and network cards. The near-empty GPU server node row is a real failure, not a plotting artifact: the class has 109 training images, a GPU server looks like a chassis from outside, and 68.4\% of the 57 test nodes are predicted as chassis, with 2.3\% correct. SXM GPU modules (49 test images) are right on 11.6\%.}
  \label{fig:confusion}
\end{figure*}
\begin{figure}[t]
  \centering
  \includegraphics[width=\linewidth]{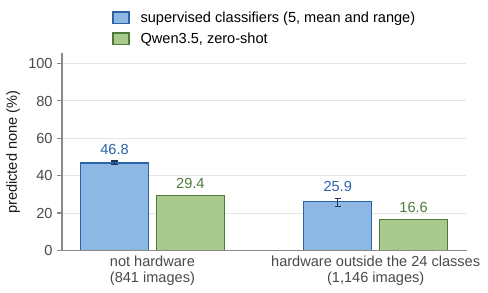}
  \caption{Share of test images human annotators marked \emph{none} that each model also calls \emph{none}, split by the annotators' granularity answer. Hardware outside the 24 classes is rejected far less often than photographs with no hardware, but even those are rejected only about half the time.}
  \label{fig:none_split}
\end{figure}
\begin{figure*}[tp]
  \centering
  \includegraphics[width=\linewidth]{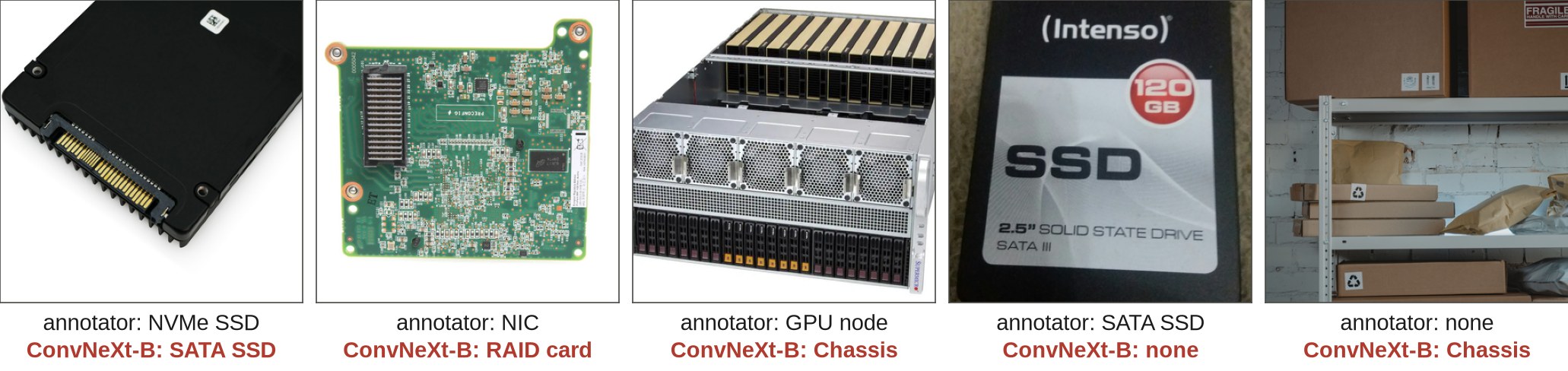}
  \caption{Typical errors of ConvNeXt-B on the human-labeled test set, with the annotators' class and the prediction under each image: look-alike drive interfaces, a network adapter on a mezzanine board, a GPU server read as a bare chassis, a part rejected as not hardware, and a shelf of boxes read as a chassis.}
  \label{fig:failures}
\end{figure*}
Rejection is the largest single source of error in the main paper, and \cref{fig:none_split} splits the 2,061 test images human annotators marked \emph{none} into 1,146 that show hardware outside the 24 classes, 841 that show no hardware and 74 that are undetermined.
The five trained classifiers reject 25.9\% of the first kind and 46.8\% of the second on average, so the taxonomy boundary explains only part of the low recall of \emph{none}.
Qwen3.5 rejects 16.6\% of the first kind and 29.4\% of the second.
\Cref{fig:failures} shows five errors of ConvNeXt-B, beside the confusion matrix of \cref{fig:confusion}.

\begin{table*}[t]
\centering
\small
\setlength{\tabcolsep}{2.5pt}
\caption{Classification on the human-labeled test images, mean $\pm$ std over seeds. Agreed: top-1 on the 782 images two different human annotators gave the same class. Named: top-1 on images human annotators gave one of the 24 classes. None: recall on images they marked none. Head, mid, tail: thirds of the classes by training frequency (none is a head class). Top box: the class of the detector's top kept box, none without one. Human annotators: one annotator scored against another on the images two annotators answered. $^\dagger$Also used to build the dataset: Qwen3.5 made the training class labels, OWLv2 and SAM~3 the object outlines and reference boxes, and the three suggestions annotators saw drew on all three and the source category. Higher is better in every column but ECE. Best in bold, top-box and human annotator rows aside. Human labels.}
\label{tab:cls_full}
\begin{adjustbox}{max width=\linewidth}
\begin{tabular}{lrrrrrrrrr}
\toprule
model & top-1 & agreed & named & none & macro-F1 & head & mid & tail & ECE \\
\midrule
\multicolumn{10}{c}{\textbf{Trained on the training set}} \\
\midrule
ResNet-50 & 63.5 $\pm$ 0.1 & 75.1 $\pm$ 0.4 & 71.6 $\pm$ 0.2 & 33.3 $\pm$ 0.4 & 60.0 $\pm$ 0.2 & 57.2 $\pm$ 0.4 & 76.8 $\pm$ 0.2 & 56.4 $\pm$ 0.3 & \textbf{6.7} $\pm$ 0.3 \\
ConvNeXt-T & 64.3 $\pm$ 0.3 & 76.9 $\pm$ 0.4 & 72.2 $\pm$ 0.3 & 34.9 $\pm$ 0.7 & 60.6 $\pm$ 0.5 & 58.2 $\pm$ 0.2 & 77.1 $\pm$ 0.3 & 57.6 $\pm$ 0.9 & 13.2 $\pm$ 0.2 \\
ConvNeXt-B & 65.0 $\pm$ 0.4 & 77.1 $\pm$ 1.0 & 73.0 $\pm$ 0.4 & 35.2 $\pm$ 0.3 & 61.1 $\pm$ 0.5 & \textbf{58.9} $\pm$ 0.4 & \textbf{77.9} $\pm$ 0.4 & 57.9 $\pm$ 0.5 & 14.1 $\pm$ 0.5 \\
ViT-B/16 & 64.6 $\pm$ 0.2 & 75.6 $\pm$ 0.4 & 72.5 $\pm$ 0.2 & 35.1 $\pm$ 0.6 & 60.9 $\pm$ 0.2 & 58.4 $\pm$ 0.8 & 77.5 $\pm$ 0.6 & \textbf{58.3} $\pm$ 0.6 & 10.7 $\pm$ 0.2 \\
EfficientNet-B3 & 64.5 $\pm$ 0.2 & 75.8 $\pm$ 0.3 & 72.0 $\pm$ 0.3 & 36.5 $\pm$ 0.5 & 60.2 $\pm$ 0.3 & \textbf{58.9} $\pm$ 0.4 & 77.0 $\pm$ 0.5 & 56.9 $\pm$ 0.6 & 11.0 $\pm$ 0.3 \\
\midrule
\multicolumn{10}{c}{\textbf{Frozen features}} \\
\midrule
DINOv2-L linear probe & 64.0 & 75.6 & 70.9 & 38.2 & 59.6 & - & - & - & 13.1 \\
DINOv2-L kNN & 63.1 & 74.7 & 70.4 & 35.8 & 58.6 & - & - & - & 13.7 \\
CLIP ViT-L/14 linear probe & 62.0 & 72.3 & 67.6 & 41.1 & 56.6 & - & - & - & 12.0 \\
CLIP ViT-L/14 kNN & 59.7 & 71.2 & 66.8 & 32.9 & 54.8 & - & - & - & 12.4 \\
CLIP ViT-B/32 linear probe & 62.0 & 73.0 & 67.4 & \textbf{42.0} & 56.7 & - & - & - & 11.7 \\
CLIP ViT-B/32 kNN & 61.4 & 72.8 & 67.8 & 37.2 & 55.9 & - & - & - & 11.8 \\
\midrule
\multicolumn{10}{c}{\textbf{Zero-shot}} \\
\midrule
CLIP ViT-L/14 & 43.2 & 52.7 & 52.4 & 9.0 & 40.1 & - & - & - & 11.0 \\
CLIP ViT-B/32 & 46.6 & 56.0 & 56.8 & 8.7 & 46.3 & - & - & - & 20.7 \\
Qwen3.5 (VLM)$^\dagger$ & \textbf{66.2} & \textbf{80.4} & \textbf{77.9} & 22.1 & \textbf{65.9} & - & - & - & - \\
Gemma~3 27B (VLM) & 59.8 & 73.3 & 70.6 & 19.2 & 59.4 & - & - & - & - \\
OWLv2, top box$^\dagger$ & 26.6 & 25.7 & 14.0 & 73.5 & 13.7 & - & - & - & - \\
SAM~3, top box$^\dagger$ & 27.5 & 30.4 & 29.9 & 18.6 & 21.6 & - & - & - & - \\
\midrule
\multicolumn{10}{c}{\textbf{Human annotators}} \\
\midrule
annotator vs. annotator & 74.0 & - & 76.8 & 62.3 & 69.0 & - & - & - & - \\
\bottomrule
\end{tabular}
\end{adjustbox}
\end{table*}

\Cref{tab:cls_full} is the full version of main-paper \cref{tab:cls}: it adds the spread over three seeds, the kNN probes, the two detector top-box rows, accuracy on head and mid classes, and calibration error.
The two top-box rows label an image with the class of the detector's top box kept at score 0.6, and \emph{none} without one: OWLv2 keeps a box on only 44.1\% of test images and so mostly answers \emph{none} (none recall 73.5\%, top-1 26.6\%), SAM~3 keeps one on 93.9\% (top-1 27.5\%), and with 14.0 and 29.9\% on named images both show that a detector's top box is a poor image label.

\section{Condition values one by one}
\label{sec:supp_cond}
\begin{table}[t]
\centering
\small
\setlength{\tabcolsep}{2.5pt}
\caption{Condition recognition, one row per condition value, on the test images. Images and share (\%): how often human annotators mark the value (an image can carry several). $\alpha$: agreement of two annotators (Krippendorff's $\alpha$, 1 is perfect, 0 is chance). Model columns: average precision (\%) of a multi-task classifier, higher is better, where a random ranking scores about the share. Last rows: mean AP over the six values and granularity accuracy (\%). Better model in bold. Human labels.}
\label{tab:cond_full}
\begin{adjustbox}{max width=\linewidth}
\begin{tabular}{lrrrrr}
\toprule
condition & images & share & $\alpha$ & ConvNeXt-T & ViT-B/16 \\
\midrule
normal & 8,012 & 82.1 & 0.23 & 87.9 & \textbf{88.3} \\
partial occlusion & 987 & 10.1 & 0.17 & 20.3 & \textbf{21.4} \\
stacked & 330 & 3.4 & 0.50 & 44.4 & \textbf{46.6} \\
broken & 71 & 0.7 & 0.11 & \textbf{10.7} & 6.6 \\
deformed & 31 & 0.3 & 0.18 & 2.7 & \textbf{3.5} \\
unusual angle & 871 & 8.9 & 0.20 & 17.9 & \textbf{19.0} \\
\midrule
mean AP &  &  &  & 30.7 & \textbf{30.9} \\
granularity acc. &  &  &  & \textbf{66.1} & \textbf{66.1} \\
\bottomrule
\end{tabular}
\end{adjustbox}
\end{table}

The main paper reports condition in three groups, and \cref{tab:cond_full} gives all six values as they are labeled and released.
The best mean AP over the six is 30.9.
Models are far above the random baseline for stacking (46.6 against 3.4) and breakage (10.7 against 0.7), about twice the baseline for partial occlusion and unusual angle, and barely above it for \emph{normal} (88.3 against 82.1).
Annotators agree on stacking ($\alpha = 0.50$) far more than on occlusion (0.17), breakage (0.11) or any other value, so part of the gap is the label and not the model.
Pooling broken and deformed into one damaged value in the main paper does not raise agreement ($\alpha$ 0.08 against 0.11 for broken alone), it only puts two rare values together.
On the 500 same-annotator repeats, broken or deformed was marked in only 3 of 996 answers and never on both copies of an image, so the same-annotator $\alpha$ of 0.00 in \cref{tab:agreement} reflects too few positives to measure.

\section{Detection with COCO AP}
\label{sec:supp_det}
\begin{table}[t]
\centering
\small
\setlength{\tabcolsep}{2.5pt}
\caption{Detection on the test images human annotators gave one of the 24 classes, all columns. Found: a detection overlaps the outlined main object at IoU 0.5. Named: it also has the human class. AP columns: COCO AP against model-made boxes on the same images, not backed by human annotators. SAM~3 has no AP, the reference boxes are kept from its own detections. Agreed boxes: trained only on the boxes both pseudo-labelers agreed on. Higher is better. $^\dagger$Made the outlines and reference boxes. Best in bold, $^\dagger$ rows aside. Human labels.}
\label{tab:det_full}
\begin{adjustbox}{max width=\linewidth}
\begin{tabular}{lrrrrr}
\toprule
detector & found & named & AP & AP50 & AP75 \\
\midrule
Faster R-CNN & 94.2 & 59.8 & 40.9 & 46.9 & 43.3 \\
RetinaNet & 96.6 & 59.9 & 41.3 & 49.3 & 43.4 \\
FCOS & 96.4 & 58.9 & \textbf{44.3} & \textbf{51.8} & \textbf{47.3} \\
YOLO11-m & 97.5 & \textbf{63.4} & 41.6 & 47.1 & 44.1 \\
Faster R-CNN (agreed boxes) & 89.5 & 30.6 & 13.4 & 16.2 & 14.2 \\
Grounding DINO (zero-shot) & \textbf{99.2} & 13.2 & 3.1 & 3.9 & 3.4 \\
OWLv2 (zero-shot)$^\dagger$ & 99.1 & 24.6 & 5.9 & 7.4 & 6.1 \\
SAM~3 (zero-shot)$^\dagger$ & 95.9 & 33.1 & - & - & - \\
\bottomrule
\end{tabular}
\end{adjustbox}
\end{table}

\begin{figure}[t]
  \centering
  \includegraphics[width=\linewidth]{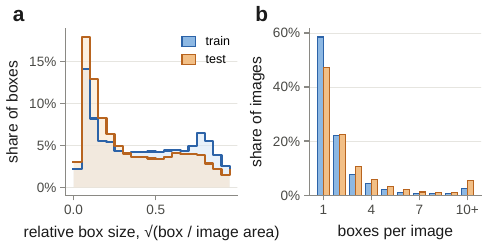}
  \caption{Are the detectors tested on the kind of boxes they were trained on? (a) The size of a box relative to its image (square root of the area share) and (b) the number of boxes per image, for the pseudo-labeled training boxes and for the test boxes. The two sets agree in shape. Test images hold somewhat more and smaller boxes (47\% of test images have a single box against 58\% of training images), because the test draw favors images on which the pseudo-labelers disagree, which are more often cluttered.}
  \label{fig:supp_boxes}
\end{figure}

\Cref{tab:det_full} adds COCO AP~\cite{lin2014coco} to the found and named scores of the main paper.
For a scored image $x$ with outline box $b_x$ and human class $c_x$, let $D_x^{+}$ be the detections that overlap the outline at $\mathrm{IoU} \ge 0.5$ and $d^\star_x$ the highest-scoring of them. With $\mathbf{1}[\cdot]$ equal to 1 when its condition holds and 0 otherwise,
\begin{equation}
  \begin{aligned}
    \mathrm{found}(x) &= \mathbf{1}\big[D_x^{+} \neq \emptyset\big],\\
    \mathrm{named}(x) &= \mathbf{1}\big[\mathrm{cls}(d^\star_x) = c_x\big],
  \end{aligned}
  \label{eq:det}
\end{equation}
where named is 0 when nothing overlaps, and both scores are means over the 7,512 test images human annotators gave a hardware class and that have an outline box.
Found counts any of the up to 100 detections a detector keeps per image, 7 to 43 on average for the four trained detectors.
With only the ten highest-scoring ones they still find the object on 93.5 to 94.9\% of images and name it on 57.7 to 62.4\%, and with the single highest-scoring one on 68.0 to 74.4\% and 44.0 to 52.0\% (\cref{tab:det_topk}).
So the object is nearly always among a detector's first ten boxes, and its first box is on it about three times in four.
Naming changes little past three boxes, which places the remaining error in the class and not in the box.
AP is measured against model-made boxes on the same test images and is not backed by human annotators, so it compares detectors with each other and with other detection work, not with the truth.
FCOS has the highest AP (44.3), the other three trained detectors lie within 0.7 points of each other (40.9 to 41.6), and the zero-shot detectors stay at 3.1 (Grounding DINO) and 5.9 (OWLv2).
SAM~3 has no AP because the reference boxes are kept from its own detections.
The OWLv2 row of this table is the zero-shot detection run prompted with one everyday name per class, a different run from the pseudo-labeling run whose top box gives the OWLv2 rows of the classification and slices tables.
One box carries one class, so an object that is two things at once, such as a GPU server node that is also a chassis, keeps only one, and training and test boxes have similar sizes and counts per image (\cref{fig:supp_boxes}).
SAM~3 zero-shot finds the outlined object on 95.9\% of images and names it on 33.1\% (main-paper \cref{tab:det}), and its saved boxes stop at score 0.4, so it is not a like-for-like detector run.

\section{Ablations and scaling}
\label{sec:supp_abl}
\begin{table}[t]
\centering
\small
\setlength{\tabcolsep}{2.5pt}
\caption{Ablations on ConvNeXt-T (change from baseline). Human labels.}
\label{tab:abl}
\begin{adjustbox}{max width=\linewidth}
\begin{tabular}{lrr}
\toprule
variant & top-1 & macro-F1 \\
\midrule
ConvNeXt-T baseline & 64.3 & 60.6 \\
near-duplicates of test images in train & 64.3 (+0.0) & 61.9 (+1.3) \\
same-page images removed from train & 63.6 (-0.7) & 59.3 (-1.2) \\
no near-duplicates in train & 63.2 (-1.1) & 58.9 (-1.7) \\
no negatives (no none class) & 57.9 (-6.4) & 56.5 (-4.0) \\
plain cross-entropy & 64.8 (+0.5) & 59.9 (-0.7) \\
384 px input & 64.6 (+0.3) & 60.7 (+0.1) \\
\bottomrule
\end{tabular}
\end{adjustbox}
\end{table}

\begin{figure}[t]
  \centering
  \includegraphics[width=\linewidth]{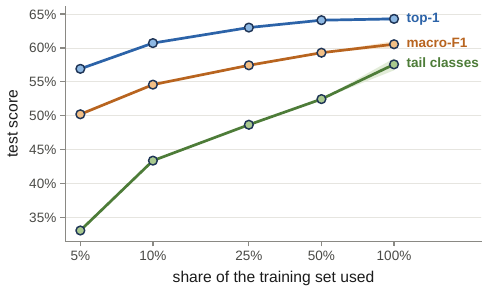}
  \caption{Does more training data still help? One classifier, ConvNeXt-T, is trained on 5, 10, 25, 50 and 100\% of the training groups and scored on the human-labeled test images: top-1, macro-F1 and accuracy on the tail classes. We use ConvNeXt-T because it is the backbone of every ablation and is within 0.7 points of the best model with a third of its parameters. The curves flatten after 25\%: more pseudo-labeled data of the same kind adds little.}
  \label{fig:scaling}
\end{figure}
\Cref{tab:abl} varies one choice at a time on ConvNeXt-T, with the test set fixed.
Putting the withheld near-duplicates of test images back into training raises top-1 by less than 0.1 points and macro-F1 by 1.3 (\cref{tab:abl}), and on the 3,588 test images that have such a near-duplicate it gives 67.6 against 67.3\%, inside the 95\% interval (\cref{tab:slices_full}).
Removing every training image that shares a source page with a test image, which near-duplicate groups do not catch, costs 0.7 points.
Dropping the non-hardware images from training costs 6.4 points, because a fifth of the test images are \emph{none}, and training on one image per near-duplicate group costs 1.1 points, so near-duplicates are worth keeping.
Plain cross-entropy gains 0.5 points of top-1 and loses 0.7 of macro-F1, and 384 px input adds 0.3.
Accuracy rises from 56.9\% with 5\% of the training set to 63.0\% with 25\% and 64.3\% with all of it (\cref{fig:scaling}), and \cref{fig:supp_labelers} compares every source of labels with the human annotators, class by class.

\section{Training settings and software}
\label{sec:supp_hparams}
Classifiers are fine-tuned from public checkpoints for 10 epochs of 200,000 sampled images each with AdamW (weight decay 0.05, one warm-up epoch, cosine decay), label smoothing 0.1 and square-root class balancing, at 224 px from 448 px cached copies.
The learning rate is set per backbone: $10^{-4}$ for ViT-B/16, $2\times10^{-4}$ for ConvNeXt-B, $4\times10^{-4}$ for ConvNeXt-T and $5\times10^{-4}$ for ResNet-50 and EfficientNet-B3, at batch 128 (96 for ConvNeXt-B).
The training pool is the 305,587 labeled training images plus 100,000 sampled non-hardware images for the \emph{none} class, and the reported checkpoint is the epoch with the best validation macro-F1.
Detectors start from COCO weights and train for 12 epochs with SGD (learning rate 0.02 for 16 images, scaled linearly), momentum 0.9 and weight decay $10^{-4}$, at 800 to 1,024 px.
Grounding DINO is prompted with one everyday name per class in a single prompt, chosen on 3,000 validation images over the class names of the annotation tool in one prompt or in three prompts of eight, and kept at the 100 highest-scoring boxes per image.
Software: PyTorch 2.13, torchvision 0.28, timm 1.0.28, transformers 5.15.1, PaddleOCR 3.7.0 (PP-OCRv5), with every model revision pinned in the released code.
Every result file records the code version, the release hash, the seed and the full configuration.

\paragraph{Seeds.} The five fine-tuned classifiers are trained with three seeds, and \cref{tab:cls_full} gives their mean and standard deviation, which is at most 0.4 points of top-1.
Every other trained model has one seed: the probes on frozen features, the multi-task condition models, the detectors, and the cross-dataset, leakage, ablation and scaling runs.
Zero-shot models have no training seed.
A difference under about one point between two single-seed rows is therefore not a ranking: seeds move top-1 by up to 0.4 points, and the 95\% interval of a top-1 near 64\% on \ntestscored test images is $\pm$1.0 point.

The classifiers are chosen so that a difference in accuracy can be traced to one factor.
ResNet-50 (ImageNet-1k weights) is the most common reference backbone, so it anchors our numbers to other work.
ConvNeXt-T and ConvNeXt-B (ImageNet-22k) are a modern convolutional design at two sizes, which shows whether capacity helps on noisy labels, and ConvNeXt-T is the backbone of every ablation because it is within 0.7 points of the best model at about a third of its parameters.
ViT-B/16 (ImageNet-21k) is the vision transformer at a pretraining scale close to that of ConvNeXt, and EfficientNet-B3 (ImageNet-1k) is meant as the kind of compact network an on-site system could run.
The five share one recipe, but their pretraining data are not identical (ImageNet-1k for two, ImageNet-21k or -22k for three), so the comparison is between public checkpoints as they are used, not between architectures in isolation.
Probes on frozen DINOv2-L, CLIP ViT-L/14 and CLIP ViT-B/32 features keep the representation fixed: DINOv2 is self-supervised and CLIP is trained on image and text pairs, so the two kinds of feature separate what the representation contributes from what our labels teach.
CLIP zero-shot is the no-training floor, and the Qwen3.5 pseudo-labeler is reported zero-shot beside it. Its labels are the training targets, and the three suggestions annotators saw, in random order beside the full list of classes, drew on Qwen3.5, SAM~3, OWLv2 and the source category, which the marks on those rows record.
Gemma 3 27B is the independent vision-language reference: it made none of our labels, no annotator saw its answers, and it was served locally with vLLM on one GPU and run zero-shot with exactly the prompt, class definitions, image size and decoding settings of the Qwen3.5 pseudo-labeling run.
Of its 9,900 answers on the test images, 8 could not be parsed and count as \emph{none}.
As in the pseudo-labeling run, the prompt carries the OCR lines for the 2,561 test images that had them in that run and ``(none)'' for the others, so the input is not uniform across images, which applies to both vision-language models alike.

The three trained detectors cover the main design families, two-stage (Faster R-CNN), anchor-based one-stage (RetinaNet) and anchor-free one-stage (FCOS).
All three come from torchvision with a ResNet-50 FPN backbone, COCO initialization, the same optimizer and the same 12-epoch schedule, so that their differences reflect the detection head and not the training setup, and the library's permissive license lets the released code carry them.
YOLO11-m is the real-time reference and is trained on the same data, classes, image size (1,024 px) and number of epochs (12) as the torchvision detectors, but with its framework's own defaults for batch size (16 against 8), augmentation (mosaic, color and scale jitter, against horizontal flip only) and optimizer settings.
It is a single untuned run, and those defaults are designed for runs of about 100 epochs, so it is a floor for what the family can do here rather than its best result.
Its test boxes are kept down to a confidence of 0.001, the usual setting for evaluating YOLO models, against 0.05 or 0.2 for the other detectors.
A lower threshold can raise AP (41.6, between RetinaNet at 41.3 and FCOS at 44.3) and the found and named rates, and at the thresholds of the other detectors YOLO11-m still names the most outlined objects of the trained detectors (62.4\% at 0.05 and 60.9\% at 0.2, against at most 59.9\% for the others).
The library is used as an external tool under its own license and is not part of the released code.
Grounding DINO and OWLv2 are two widely used open-vocabulary detectors, evaluated without training to test whether web-scale grounding data cover the names of data-center parts.

The metrics follow from how the test set was drawn and labeled.
Top-1 over all images is the headline number, and it is split into top-1 on the images human annotators named and the recall of \emph{none}, because the two errors have different causes: confusing two parts, and failing to reject an image that shows no part.
The test draw is balanced by class while training is long-tailed, so macro-F1 and head, mid and tail accuracy keep the rare classes visible, and expected calibration error is reported because a triage decision would act on the model's confidence.
Condition is multi-label and its non-normal values are rare, so each value is scored by average precision, which needs no decision threshold and whose chance level is the value's share.
Detection is scored by whether the outlined main object is found and named, because the annotators gave the class of that object and checked no box, and COCO AP on model-made boxes is kept for comparison with other detection work.
Text is scored by whether a reader's output contains the identifier an annotator typed, because readers return every line on the image while annotators typed only the identifying text.
Training uses an inpainted copy~\cite{suvorov2022lama} for about 23,000 images with a reseller overlay (main-paper \cref{sec:labels}), because an overlay would let a classifier read the seller in place of the part.
Validation and test images are never modified.

\section{Full audit of the public collections}
\label{sec:supp_audit}
\begin{table}[t]
\centering
\small
\setlength{\tabcolsep}{2.5pt}
\caption{Full audit of the nine public e-waste collections. Agree and box (\%) compare model-made readings, so they measure disagreement, not error. A dash marks no mapped label or no boxes.}
\label{tab:g4_full}
\begin{adjustbox}{max width=\linewidth}
\begin{tabular}{lrrrrrr}
\toprule
collection & images & copies & shared & DC & agree & box \\
\midrule
Kaggle E-Waste~\cite{akshat2023ewaste} & 3,000 & 36 & 146 & 0 & - & - \\
PC Parts Det.~\cite{cas2024pcparts} & 978 & 570 & 0 & 701 & 90.2 & 69.1 \\
EBS E-Waste~\cite{ebs2025ewaste} & 4,627 & 3,248 & 505 & 346 & 81.2 & 74.4 \\
EWasteNet~\cite{islam2023ewastenet} & 881 & 15 & 28 & 0 & - & - \\
Kaggle PC Parts~\cite{asaniczka2024pcparts} & 3,279 & 203 & 32 & 957 & 84.8 & - \\
E-Waste Cls.~\cite{manoj2024ewaste} & 19,130 & 2,188 & 19,122 & 441 & 86.0 & - \\
Laptop Comp.~\cite{islam2024laptop} & 29,120 & 25,480 & 831 & 5,600 & 71.1 & - \\
WEEE, balanced~\cite{ewd2024r0ojc} & 7,211 & 1,278 & 6,880 & 380 & 81.3 & 92.2 \\
WEEE~\cite{ewd2024r0ojc} & 19,604 & 3,486 & 19,129 & 441 & 85.3 & 81.9 \\
\midrule
all nine & 87,830 & 36,504 & 39,437 & 8,866 &  &  \\
\bottomrule
\end{tabular}
\end{adjustbox}
\end{table}

\Cref{tab:g4_full} repeats the audit of main-paper \cref{tab:g4} with two more columns: \emph{agree} is the share of images whose native label maps to one of our classes and matches an independent vision-language reading, and \emph{box} is the share of native boxes matched by a detector box at IoU 0.5.
Both compare model-made readings, so they measure disagreement, not error.

\section{Comparison with general waste datasets}
\label{sec:supp_waste}
\begin{table}[t]
\centering
\small
\setlength{\tabcolsep}{3pt}
\caption{\ewn against general waste-recognition datasets. Domains: household waste (TrashNet), litter (TACO), hand-held waste (MJU-Waste), a recycling conveyor (ZeroWaste-f), consumer e-waste (the nine public e-waste collections of main-paper \cref{tab:g4}) and data-center hardware (\ewn). Cond., text, gran.\ and boxes say whether condition, printed text, more than one granularity level and object boxes are annotated, as each dataset's own paper or page reports them, with instance masks counted as boxes. Images are unique images for the nine public e-waste collections after our deduplication. $^\dagger$Model-made boxes, none verified by an annotator.}
\label{tab:compare}
\begin{adjustbox}{max width=\linewidth}
\begin{tabular}{lrrcccc}
\toprule
dataset & images & classes & cond. & text & gran. & boxes \\
\midrule
TrashNet~\cite{yang2016trashnet} & 2,527 & 6 & \xmark & \xmark & \xmark & \xmark \\
TACO~\cite{proenca2020taco} & 1,500 & 60 & \xmark & \xmark & \cmark & \cmark \\
MJU-Waste~\cite{wang2020mjuwaste} & 2,475 & 1 & \xmark & \xmark & \xmark & \cmark \\
ZeroWaste-f~\cite{bashkirova2022zerowaste} & 4,503 & 4 & \xmark & \xmark & \xmark & \cmark \\
Nine e-waste sets & 79,738 & varies & \xmark & \xmark & \xmark & 4 of 9 \\
\midrule
\textbf{\ewn} & \textbf{328,239} & 24 & \cmark & \cmark & \cmark & \cmark$^\dagger$ \\
\bottomrule
\end{tabular}
\end{adjustbox}
\end{table}
\Cref{tab:compare} sets \ewn beside four general waste datasets that the main paper discusses.
None of them annotates condition or printed text, and only TACO, with its 28 super-categories, has more than one granularity level.

\section{Tables behind the figures}
\label{sec:supp_tables}
\begin{figure}[t]
  \centering
  \includegraphics[width=\linewidth]{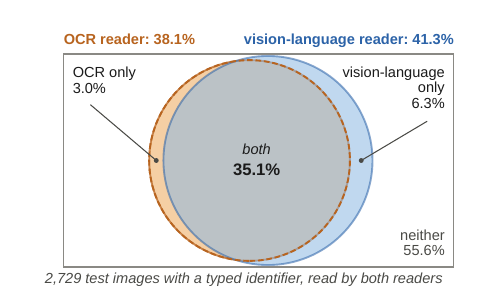}
  \caption{Text perception: test images with a typed identifier that each reader contains, over the 2,729 images both readers read (case and spacing ignored): both 35.1\%, OCR only 3.0\%, vision-language only 6.3\%, neither 55.6\%. Where the vision-language reader gave no reading, the OCR line stands in for it, so the overlap is an upper bound.}
  \label{fig:text}
\end{figure}

Three results are drawn as figures, and the tables below give their exact numbers.
\Cref{tab:funnel} is the composition of the corpus behind main-paper \cref{fig:composition}, \cref{tab:text} the text statistics and reader scores behind \cref{fig:text}, and \cref{tab:cross} the transfer results behind main-paper \cref{fig:cross}.
\begin{table}[t]
\centering
\small
\caption{Composition of the corpus after exact deduplication. The benchmark is the first three parts, 328,239 labeled hardware images.}
\label{tab:funnel}
\begin{adjustbox}{max width=\linewidth}
\begin{tabular}{lr}
\toprule
 & images \\
\midrule
unique images & 762,129 \\
near-duplicate groups & 389,169 \\
\midrule
train & 305,587 \\
validation & 12,652 \\
test & 10,000 \\
near-duplicates of test images, withheld & 10,955 \\
negatives: photographs of other objects & 370,665 \\
renders and graphics (relevance filter only) & 48,866 \\
hardware with no usable class & 3,404 \\
\bottomrule
\end{tabular}
\end{adjustbox}
\end{table}

\begin{table}[t]
\centering
\small
\caption{Text perception. Top: text statistics and the agreement of the OCR and vision-language readers per line. Bottom: a linear probe on frozen CLIP ViT-L/14 features, with and without the OCR text, on images that carry text. Human labels.}
\label{tab:text}
\begin{adjustbox}{max width=\linewidth}
\begin{tabular}{lr}
\toprule
 &  \\
\midrule
images with text & 54.3\% \\
lines per image with text & 10.1 \\
two readers, exact match & 47.2\% \\
two readers, median CER & 0.161 \\
\midrule
image features, top-1 & 60.9 \\
image and OCR text, top-1 & 62.3 \\
\midrule
OCR reader contains the typed identifier & 35.3\% \\
vision-language reader contains the typed identifier & 41.3\% \\
\bottomrule
\end{tabular}
\end{adjustbox}
\end{table}

\begin{table}[t]
\centering
\small
\setlength{\tabcolsep}{2.5pt}
\caption{Cross-dataset transfer: ConvNeXt-T trained on the nine public collections only, on the images E-Waste-Bench collected only, or on all E-Waste-Bench training images, and tested on the human-labeled test images. Gain: top-1 points over training on the public collections only. Higher is better. Best in bold. Human labels.}
\label{tab:cross}
\begin{adjustbox}{max width=\linewidth}
\begin{tabular}{lrrrr}
\toprule
training data & top-1 & named & macro-F1 & gain \\
\midrule
public only & 31.5 & 15.5 & 16.6 & - \\
collected only & 64.2 & \textbf{72.3} & \textbf{60.7} & +32.7 \\
all training images & \textbf{64.3} $\pm$ 0.3 & 72.2 $\pm$ 0.3 & 60.6 $\pm$ 0.5 & +32.8 \\
\bottomrule
\end{tabular}
\end{adjustbox}
\end{table}

\begin{figure}[t]
  \centering
  \includegraphics[width=\linewidth]{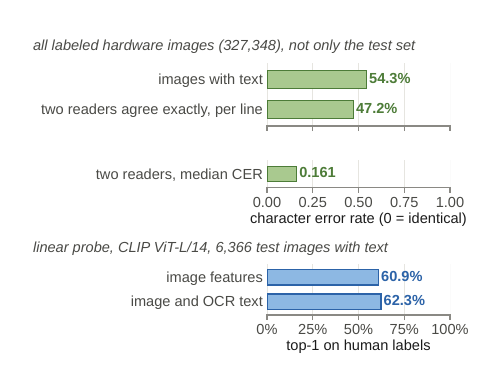}
  \caption{Text statistics over the labeled hardware images: share with text, exact agreement of the two readers per line, and their median character error rate.}
  \label{fig:text_stats}
\end{figure}

\Cref{fig:text_stats} draws the text statistics that main-paper \cref{sec:text} quotes, \cref{fig:text} the overlap of the two readers on the typed identifiers, and \cref{tab:slices_full} gives the slices of the test set that main-paper \cref{sec:cls} quotes, with the spread over three seeds.
Its OWLv2 and SAM~3 top-box columns, shown for reference because these models made the outlines, stay between about 20 and 40\% on most slices.
\begin{table}[t]
\centering
\small
\setlength{\tabcolsep}{2pt}
\caption{Top-1 of ConvNeXt-T on slices of the test set, with the spread over three seeds. Grouped: our split by near-duplicate group. Pre-labels: the pseudo-labels the test draw used. Leaky: the same model trained with the withheld near-duplicates of test images put back into training. OWLv2 and SAM~3: the class of their top kept box, as in the classification table. $^\dagger$Made the object outlines. Blocks: how the image was drawn, reseller overlay, near-duplicates, source type. Human labels.}
\label{tab:slices_full}
\begin{adjustbox}{max width=\linewidth}
\begin{tabular}{lrrrrr}
\toprule
 &  & \multicolumn{4}{c}{top-1} \\
\cmidrule(lr){3-6}
test images & n & grouped & leaky & OWLv2$^\dagger$ & SAM~3$^\dagger$ \\
\midrule
pre-labels agree & 7,583 & 65.9$\pm$0.2 & 66.3 & 22.7 & 27.5 \\
pre-labels disagree & 2,178 & 58.9$\pm$0.5 & 57.5 & 40.1 & 27.6 \\
\midrule
no overlay & 8,370 & 64.6$\pm$0.3 & 64.4 & 26.8 & 28.2 \\
reseller overlay & 1,391 & 62.5$\pm$0.1 & 63.7 & 25.0 & 23.3 \\
\midrule
no near-duplicate & 6,173 & 62.6$\pm$0.4 & 62.4 & 30.6 & 28.5 \\
near-dup. withheld & 3,588 & 67.3$\pm$0.2 & 67.6 & 19.6 & 25.9 \\
\midrule
discovered sources & 6,203 & 64.0$\pm$0.2 & 64.2 & 21.6 & 26.4 \\
archives & 2,388 & 59.7$\pm$0.3 & 59.3 & 41.4 & 32.7 \\
refurbishers & 689 & 79.8$\pm$0.8 & 78.8 & 4.9 & 14.8 \\
product catalog & 297 & 84.7$\pm$0.6 & 84.8 & 54.5 & 42.8 \\
marketplace, auction & 178 & 42.1$\pm$1.4 & 46.1 & 39.9 & 21.9 \\
\bottomrule
\end{tabular}
\end{adjustbox}
\end{table}

On the 2,506 of the 2,729 images of \cref{fig:text} for which Gemma 3 also returned text, its independent reading contains the identifier on 37.6\%, against 41.7\% for Qwen3.5 and 38.5\% for the OCR reader alone.

\section{More statistics}
\label{sec:supp_figs}
\begin{figure*}[tp]
  \centering
  \includegraphics[width=\linewidth]{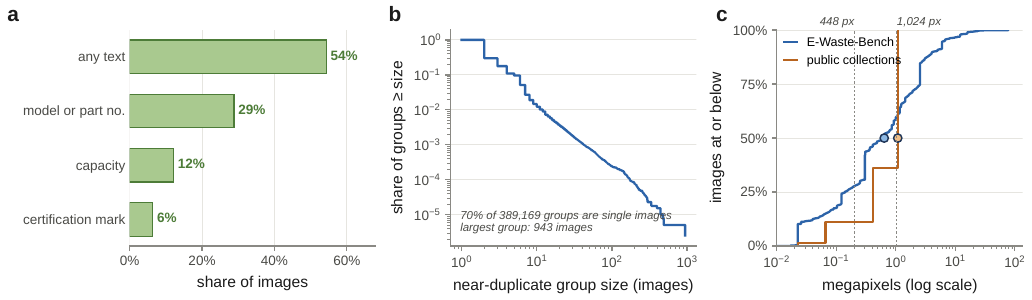}
  \caption{Statistics the main text quotes as numbers. (a) Text evidence (\gap{3}): the share of labeled images on which the readers find any text, a model or part number, a capacity, or a certification mark. (b) Near-duplicate groups: the share of groups that hold at least the given number of images, on log axes. Of the 389,169 groups, 70\% are single images and the largest holds 943. The groups, not the images, are the unit of the splits, which is why this distribution matters: a random image split would cut through every group to the right of 1. (c) Image resolution: the share of images at or below a given size, with a dot at each median and dashed lines at the 448 and 1,024 px training sizes. The public collections are resized exports, so half of their images have one single size and none exceeds 1.1 megapixels, while \ewn spans three orders of magnitude, from thumbnails to originals of several megapixels.}
  \label{fig:supp_stats}
\end{figure*}
\begin{figure}[t]
  \centering
  \includegraphics[width=\linewidth]{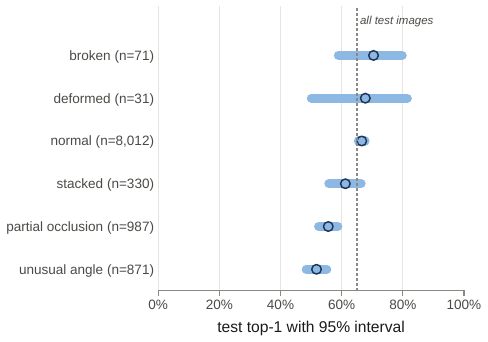}
  \caption{Does the condition of a part make it harder to classify? Top-1 of the best classifier on the test images, grouped by the condition human annotators gave them, with 95\% Wilson intervals~\cite{wilson1927interval} and the number of images $n$. The dashed line is all test images. Parts seen from an unusual angle or partly occluded are 15 and 11 points harder than parts the annotators call normal. Broken and deformed parts are not harder, but only 71 and 31 test images carry those labels, so their intervals are wide.}
  \label{fig:supp_cond}
\end{figure}

\begin{figure*}[tp]
  \centering
  \includegraphics[width=\linewidth]{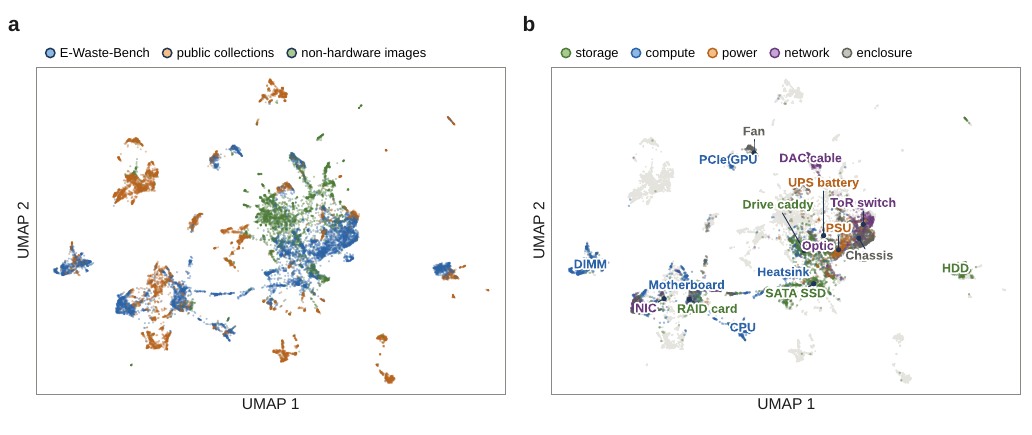}
  \caption{A map of image features: UMAP~\cite{mcinnes2018umap} of DINOv2-L~\cite{oquab2024dinov2} features for a balanced sample of 20,000 images (8,000 from \ewn, 8,000 from the public collections, 4,000 non-hardware images). The axes are the two UMAP coordinates and have no unit. (a) By origin. The public collections occupy regions that \ewn barely enters: in feature space only 7\% of a public image's ten nearest neighbors are \ewn images, where 40\% would be expected if the two were mixed, and non-hardware images keep to themselves (74\% of their neighbors are non-hardware). (b) \ewn images by function group, with each class that forms a compact cluster named at its center. Classes here are model-made labels. Two thirds of an image's neighbors share its class (67\%, against 6\% at random). Network cards and RAID cards have the closest centers of all class pairs, which matches their confusion in \cref{fig:confusion}. Risers, NVMe drives, PDUs, RAID batteries, backplanes and the two rarest GPU classes form no cluster of their own and are not named.}
  \label{fig:supp_umap}
\end{figure*}
Collected images have a median of 0.64 megapixels against 1.08 for the public collections, which were exported at one size, but cover a far wider range, 5\% under 0.02 and 5\% above 6.7 megapixels (\cref{fig:supp_stats}c).
Images with text carry 10 lines on average (\cref{fig:supp_stats}a).
In feature space the collected images and the public collections barely mix (\cref{fig:supp_umap}), the gap that the cross-dataset transfer of main-paper \cref{sec:cross} measures.

\section{Privacy, licensing and ethics in full}
\label{sec:supp_ethics}
\ewn depicts hardware, and no personal profiles or account data were collected.
People still appear incidentally, so we ran a privacy audit with SAM~3~\cite{carion2025sam3} prompted for faces, people and license plates over training, validation and test.
It flagged 3,557 of the 328,239 usable images (1.1\%), 404 of them test images, on 188 of which the automatic blur was applied.
The released copies of those test images have the detected regions blurred, and the same pass runs on training and validation before public release.
Two co-authors then checked all 404 flagged test images by hand, one reviewer per image.
Of the 188 blurred images, the blur was correct and complete on 169, covered a printed product label taken for a license plate on 10 (10 of the 12 plate detections), and was incomplete on 9, and on 6 further images an identifiable face had not been blurred at all.
Precision counts only the 10 blurs over a product label as wrong (169 of 179, 94.4\%), and recall counts the 9 incomplete and 6 missing blurs as misses (169 of 184, 91.8\%).
Only flagged images were reviewed, so this recall is an upper bound for the whole test set.
For the release, 11 of the 15 images with a missed or incomplete blur are re-blurred by hand and 4 are withheld, so the released test set is at most 4 images smaller than the scored one, the blur is removed from the 10 product labels, since it erases text evidence, and plate detections are reviewed by hand rather than blurred automatically.
The test set was labeled by graduate students in electrical and computer engineering who are familiar with data-center hardware. They were paid for the work and were trained with a tutorial and a detailed written manual, whose label definitions are given in \cref{sec:supp_annotation}. No personal data about them was collected, and their answers and answer times are reported per anonymous account.
We submitted the annotation protocol to our institution's review board, which determined that it is not human-subjects research.

Every image keeps its source URL, collection time and the source policy in force, and licensing is recorded per image.
A full licensing, IRB and legal review of the collected images follows collection (\cref{sec:supp_collection}), and images whose license does not permit redistribution are excluded at release export, with their provenance kept so that they can be retrieved for research under the original terms.
Documented removal requests are honored through a logged removal process that also blocks the source from future collection.
Watermark-inpainted copies of training images are research artifacts and are not released before the legal review, because removing a watermark removes copyright-management information.
Beyond licensing, recognizing hardware and reading its identifiers could also help locate equipment for theft or to target drives that still hold data.
The dataset contains no storage contents, and we release it for sorting, reuse and recycling research: the annotations are under CC BY-NC 4.0, and access requires accepting terms of use that prohibit using the data to locate equipment for theft or to recover data from storage devices.

\section{Privacy examples}
\label{sec:supp_privacy}
\begin{figure*}[tp]
  \centering
  \includegraphics[width=\linewidth]{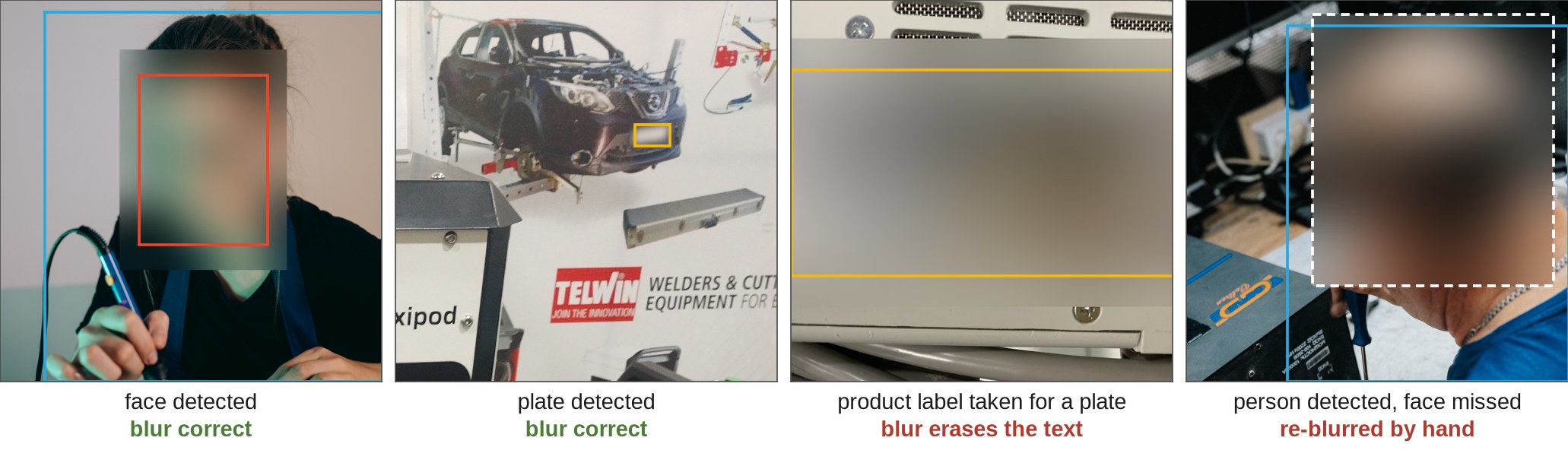}
  \caption{Four test images from the manual check of the automatic blur (\cref{sec:supp_ethics}). The boxes are the audit's detections: face (red), license plate (yellow) and person (blue). From left: a face and a plate that were found and blurred correctly, a printed label on a chassis that was taken for a plate, so that the blur erases its text, and a face seen from the side that the audit missed, shown here after its re-blur with a box set by hand (white, dashed).}
  \label{fig:privacy}
\end{figure*}
\Cref{fig:privacy} shows the automatic blur at work and its two failure modes, one example of each from the manual check.

\section{Image credits}
\label{sec:supp_credits}
The DIMMs (top row, second) and the opened hard drive (bottom row, third) of \cref{fig:overview} are under the Pexels and Unsplash licenses, which require no attribution, as are the first and fourth images of \cref{fig:privacy} (Pexels license).
{\let\scriptsize\relax
\par\begingroup\scriptsize
\noindent All 24 images of \cref{fig:label_axes} need no attribution (Pexels or Unsplash license, CC0 or public domain).
\endgroup

\par\begingroup\scriptsize
\noindent The gallery images are shown as small crops to illustrate the classes and remain the property of their owners.
\endgroup

\par}

\section{Detection with fewer boxes per image}
\label{sec:supp_det_topk}
\begin{table}[t]
\centering
\small
\setlength{\tabcolsep}{3pt}
\caption{Found and named when only the $k$ highest-scoring detections of an image count, in percent. Boxes: mean detections kept per image. Agreed boxes: trained only on the boxes both pseudo-labelers agreed on. $k = 100$ reproduces main-paper \cref{tab:det} to within 0.1 point. $^\dagger$Made the outlines and reference boxes.}
\label{tab:det_topk}
\begin{adjustbox}{max width=\linewidth}
\begin{tabular}{lrrrrr}
\toprule
 & & \multicolumn{4}{c}{$k$} \\
\cmidrule(lr){3-6}
detector & boxes & 1 & 3 & 10 & 100 \\
\midrule
\multicolumn{6}{c}{found (\%)} \\
\midrule
Faster R-CNN & 6.6 & 74.1 & 90.2 & 93.6 & 94.2 \\
RetinaNet & 43.4 & 74.2 & 88.2 & 93.5 & 96.6 \\
FCOS & 25.3 & 68.0 & 86.9 & 93.7 & 96.4 \\
YOLO11-m & 22.5 & 74.4 & 89.6 & 94.9 & 97.5 \\
Faster R-CNN (agreed boxes) & 4.0 & 78.5 & 87.0 & 89.2 & 89.5 \\
Grounding DINO (zero-shot) & 73.8 & 56.7 & 80.9 & 94.2 & 99.2 \\
OWLv2 (zero-shot)$^\dagger$ & 47.8 & 73.7 & 88.9 & 96.2 & 99.0 \\
SAM~3 (zero-shot)$^\dagger$ & 12.9 & 82.8 & 89.4 & 94.6 & 95.9 \\
\midrule
\multicolumn{6}{c}{named (\%)} \\
\midrule
Faster R-CNN & & 49.7 & 57.8 & 59.6 & 59.8 \\
RetinaNet & & 48.9 & 55.9 & 58.6 & 59.8 \\
FCOS & & 44.0 & 54.0 & 57.7 & 58.9 \\
YOLO11-m & & 52.0 & 59.8 & 62.4 & 63.4 \\
Faster R-CNN (agreed boxes) & & 26.3 & 29.3 & 30.5 & 30.6 \\
Grounding DINO (zero-shot) & & 9.6 & 12.2 & 13.0 & 13.2 \\
OWLv2 (zero-shot)$^\dagger$ & & 18.8 & 22.5 & 24.1 & 24.6 \\
SAM~3 (zero-shot)$^\dagger$ & & 27.6 & 30.1 & 32.5 & 33.1 \\
\bottomrule
\end{tabular}
\end{adjustbox}
\end{table}
\Cref{tab:det_topk} is the table behind the last paragraph of \cref{sec:supp_det}: found and named when only the highest-scoring detections of an image count.

\section{Did the suggestions steer the annotators?}
\label{sec:supp_anchoring}
\begin{table}[t]
\centering
\small
\setlength{\tabcolsep}{2.5pt}
\caption{Where the annotators' first answers fall among the three suggested classes they saw, in percent of answers, and the class agreement of two annotators on doubly answered images (number of pairs). 1st to 3rd: the rank of the suggestion by the model scores, not its place on screen, where the three were shown in random order. Agreed and disputed: whether the pseudo-labelers agreed on the image when it was drawn. The other 98 answers, on 78 images drawn where Qwen3.5 agreed with the source category but no detector did, count only in the all row.}
\label{tab:anchoring}
\begin{adjustbox}{max width=\linewidth}
\begin{tabular}{lrrrrrrr}
\toprule
 & & \multicolumn{3}{c}{suggested class} & & & \\
\cmidrule(lr){3-5}
images & answers & 1st & 2nd & 3rd & other & none & agree (pairs) \\
\midrule
all & 11,457 & 61.7 & 7.0 & 4.9 & 6.3 & 20.0 & 74.0 (1,057) \\
agreed & 8,812 & 69.3 & 6.3 & 4.8 & 6.2 & 13.4 & 74.4 (825) \\
disputed & 2,547 & 36.0 & 9.7 & 5.7 & 6.7 & 41.9 & 73.6 (220) \\
\bottomrule
\end{tabular}
\end{adjustbox}
\end{table}

Annotators saw three suggested classes with every image (main-paper \cref{sec:labels}), and the class of Qwen3.5 was the first suggestion on 90.0\% of the 9,900 images the annotators saw.
No image was annotated without suggestions, so their effect cannot be measured directly, and \cref{tab:anchoring} bounds it instead.
First, annotators did not simply accept the first suggestion: 61.7\% of the 11,457 first answers are suggestion 1, and 26.3\% are \emph{none} or a class that was not offered.
The share of suggestion 1 runs from 55.7 to 68.7\% across the eleven accounts.
Second, where the pseudo-labelers disputed an image when it was drawn, annotators took the first suggestion on 36.0\% of answers, against 69.3\% where they agreed.
Third, two annotators agree as often on disputed images (73.6\% of 220 pairs) as on agreed ones (74.4\% of 825), which argues against agreement produced by a shared suggestion.
Fourth, Gemma~3, whose answers no annotator saw, disagrees with the first suggestion on 3,429 test images that have a human class.
There the human class is the first suggestion on 29.6\%, the class of Gemma~3 on 25.0\% and neither on 45.4\%.
The shown suggestion thus wins 54\% of the conflicts that one of the two wins, an edge that mixes anchoring with Qwen3.5 being the stronger model (main-paper \cref{tab:cls}).
These numbers bound the effect of the suggestions and do not rule it out.

\section{Per-class reliability of the test set}
\label{sec:supp_class_reliability}
\begin{table}[t]
\centering
\small
\setlength{\tabcolsep}{2.5pt}
\caption{Reliability of each class row of the human-labeled test set. n: scored test images of the class. Agree: share of pairs of answers from two annotator accounts that give the same class, on images of the class (number of pairs). Recall: ConvNeXt-B on the images of the class, mean over three seeds, with a 95\% Wilson interval over images. Wide: the recall (R) or agreement (A) interval is wider than 20 points, which holds for 0 and 18 of the 24 classes. No class: images still tied after the third votes, not scored, their pairs count in the all row.}
\label{tab:class_reliability}
\begin{adjustbox}{max width=\linewidth}
\begin{tabular}{lrrrrl}
\toprule
class & n & agree (pairs) & recall & 95\% interval & wide \\
\midrule
HDD & 447 & 89.7 (29) & 86.6 & 83.1 to 89.4 & A \\
SATA SSD & 446 & 80.7 (62) & 75.0 & 70.7 to 78.7 &  \\
NVMe SSD & 333 & 86.2 (29) & 67.4 & 62.1 to 72.1 & A \\
Drive caddy & 236 & 62.5 (24) & 72.7 & 66.9 to 78.1 & A \\
Backplane & 221 & 72.1 (43) & 32.7 & 26.7 to 39.0 & A \\
RAID card & 258 & 53.3 (15) & 62.4 & 56.4 to 68.1 & A \\
RAID battery & 199 & 91.7 (12) & 65.0 & 58.0 to 71.1 & A \\
CPU & 461 & 97.0 (67) & 95.0 & 92.6 to 96.7 &  \\
DIMM & 416 & 100.0 (31) & 90.1 & 86.9 to 92.7 &  \\
Motherboard & 573 & 75.0 (40) & 60.3 & 56.3 to 64.3 & A \\
PCIe GPU & 365 & 88.9 (27) & 87.3 & 83.6 to 90.4 & A \\
SXM GPU & 49 & 50.0 (4) & 11.6 & 5.7 to 24.2 & A \\
Heatsink & 363 & 89.1 (46) & 84.5 & 80.5 to 87.9 &  \\
Riser & 233 & 88.9 (18) & 76.0 & 70.1 to 81.0 & A \\
PSU & 484 & 89.7 (87) & 69.4 & 65.2 to 73.4 &  \\
UPS battery & 179 & 76.5 (17) & 43.4 & 36.5 to 50.9 & A \\
PDU & 133 & 92.3 (13) & 38.6 & 30.5 to 46.8 & A \\
NIC & 498 & 87.1 (31) & 67.0 & 62.8 to 71.0 & A \\
Optic & 312 & 87.5 (24) & 81.7 & 77.1 to 85.6 & A \\
DAC cable & 202 & 35.3 (17) & 81.8 & 75.8 to 86.4 & A \\
ToR switch & 361 & 81.5 (27) & 78.0 & 73.6 to 82.1 & A \\
Chassis & 431 & 68.4 (38) & 82.6 & 78.7 to 85.9 & A \\
GPU node & 57 & 0.0 (1) & 2.3 & 0.3 to 9.3 & A \\
Fan & 443 & 96.1 (77) & 70.9 & 66.5 to 74.9 &  \\
\midrule
none & 2,061 & 77.6 (165) & 35.2 & 33.1 to 37.3 &  \\
no class & - & 0.0 (113) & - & - &  \\
all & 9,761 & 74.0 (1,057) & 65.0 & 64.0 to 65.9 &  \\
\bottomrule
\end{tabular}
\end{adjustbox}
\end{table}

\Cref{tab:class_reliability} asks, class by class, how far a row of the benchmark can be trusted: how many scored test images the class has, how often two annotators give it the same class, and how wide the interval of the best classifier's recall is.
Recall is measured well enough to compare models in every class.
Its 95\% interval is 4 to 16 points wide for 22 classes and widest for SXM GPU modules (5.7 to 24.2\% on 49 images), which with GPU nodes (57 images) is one of the two classes that are both small and nearly unrecognized.
Agreement per class is only indicative, because a class has between 1 and 87 doubly answered images and the interval is wider than 20 points for 18 of the 24 classes.
Where the pairs are many the picture is clear: annotators agree on CPUs (97.0\% of 67 pairs), fans (96.1\% of 77) and power supplies (89.7\% of 87), and less on SATA SSDs (80.7\% of 62), motherboards (75.0\% of 40), backplanes (72.1\% of 43) and chassis (68.4\% of 38).
Direct-attach cables stand out the other way, with agreement on 35.3\% of 17 pairs beside a recall of 81.8\%, so there the open question is the class boundary and not the classifier.
The agreement of a class is above the pooled 74.0\% in most rows because the 113 doubly answered images that stayed tied have no class and count only in the last row.

\section{Accuracy and the training exposure of a source}
\label{sec:supp_source_exposure}
\begin{table}[t]
\centering
\small
\setlength{\tabcolsep}{3pt}
\caption{Top-1 of ConvNeXt-T (mean $\pm$ std over three seeds) on test images grouped by how many labeled training images come from the same source host. Src: distinct sources. In brackets under top-1: its 95\% interval from resampling sources. None: percent of images whose human class is none. Named: top-1 on images whose human class is one of the 24. Mix: the top-1 the bucket's class mix predicts from the other buckets. 4 of the 9,761 scored test images, from one source, have no training image from their source.}
\label{tab:source_exposure}
\begin{adjustbox}{max width=\linewidth}
\begin{tabular}{lrrrrrr}
\toprule
train images & n & src & none & top-1 & named & mix \\
\midrule
0 to 99 & 355 & 22 & 28.2 & 60.3 $\pm$ 1.4 & 71.5 & 60.7 \\
 & & & & [40.0, 73.3] & & \\
100 to 999 & 2,559 & 40 & 22.0 & 71.1 $\pm$ 0.4 & 76.0 & 61.9 \\
 & & & & [65.7, 77.2] & & \\
1,000 to 9,999 & 2,208 & 12 & 19.1 & 61.7 $\pm$ 0.5 & 69.4 & 63.5 \\
 & & & & [53.0, 69.7] & & \\
10,000 or more & 4,639 & 6 & 21.0 & 62.0 $\pm$ 0.1 & 71.5 & 66.1 \\
 & & & & [58.5, 70.9] & & \\
\midrule
all & 9,761 & 80 & 21.1 & 64.3 $\pm$ 0.2 & 72.2 & \\
\bottomrule
\end{tabular}
\end{adjustbox}
\end{table}

The test set cannot measure generalization to unseen web sources: only 4 of the \ntestscored scored test images come from a source without a labeled training image.
What it can measure is whether accuracy follows the amount a source contributed to training.
\Cref{tab:source_exposure} groups the test images by the number of labeled training images from the same source, and its last column gives the top-1 each group would reach if only its mix of classes mattered, computed from the accuracy of every class on the images outside the group.
There is no trend.
The 22 least exposed sources, with under 100 training images each, reach 60.3\% where their class mix predicts 60.7\%, and the six sources with 10,000 or more training images reach 62.0\% against a predicted 66.1\%.
Only the group with 100 to 999 training images is above its prediction (71.1\% against 61.9\%), and it holds all 689 refurbisher photographs, which score 79.8\% (\cref{tab:slices_full}).
Accuracy therefore follows the kind of source and not how much of that source the classifier saw.
The intervals, from resampling sources, are wide (40.0 to 73.3\% for the least exposed group), exposure is counted per host, and the kind of source is not separated from exposure, so this is evidence against memorized sources and not a test on new ones.

\section{Model-made against human labels in training}
\label{sec:supp_foldft}
\begin{table}[t]
\centering
\small
\setlength{\tabcolsep}{2.5pt}
\caption{Five-fold cross-validated fine-tuning of the seed-0 ConvNeXt-T on the scored test images (9,761 images, fold split by human class and near-duplicate group). Each image is predicted by a model fine-tuned on the other four folds and never on its own label, and is scored against its human label. Columns: the unchanged model, trained on the release training labels, the model fine-tuned on model-made or on human labels, and the model with only its head fine-tuned on human labels. Model-made labels are the pseudo-labels the same images carry in the release. The fine-tuned models have seen test-distribution images, so this measures label quality and is not a benchmark result. Top-1, named, none and macro-F1 in percent, with the 95\% bootstrap interval over images in brackets on the line below. Change: change in top-1 against the unchanged model.}
\label{tab:foldft}
\begin{adjustbox}{max width=\linewidth}
\begin{tabular}{lrrrr}
\toprule
 & unchanged & \multicolumn{2}{c}{fine-tuned} & head only \\
\cmidrule(lr){3-4}
labels & release & model-made & human & human \\
\midrule
top-1 & 64.1 & 62.6 & 68.9 & 67.3 \\
 & [63.1, 65.0] & [61.7, 63.6] & [67.9, 69.8] & [66.3, 68.3] \\
named & 72.1 & 73.8 & 75.3 & 74.9 \\
none & 34.1 & 21.0 & 44.8 & 38.8 \\
macro-F1 & 60.0 & 61.4 & 66.2 & 64.8 \\
 & [59.0, 61.0] & [60.4, 62.4] & [65.1, 67.3] & [63.7, 65.8] \\
\midrule
change &  & $-$1.4 & +4.8 & +3.2 \\
 &  & [$-$2.2, $-$0.7] & [4.1, 5.5] & [2.7, 3.8] \\
\bottomrule
\end{tabular}
\end{adjustbox}
\end{table}

\paragraph{What \emph{none} is on each side.}
The training \emph{none} class and the test \emph{none} class are different kinds of image.
Training draws 100,000 of 370,665 negatives: 303,534 photographs that the pseudo-labeler read as showing no data-center hardware (114,483 of them by sharing the reading of a near-duplicate), 63,933 images of the public collections whose own label is an electronic device outside the 24 classes, and 3,198 set aside at collection time.
Of the 2,061 test images that annotators marked \emph{none}, 1,605 (78\%) are images the pseudo-labeler had given one of the 24 classes.
No such image can be a training negative, since an image the pseudo-labeler takes for a class is trained as that class.
The classifiers therefore learn to reject what does not look like data-center hardware and are tested mostly on what does, which is where rejection is weakest (\cref{fig:none_split}).

\paragraph{What human labels buy.}
\Cref{tab:foldft} measures the value of cleaner labels directly.
The scored test images are split into five folds by class and near-duplicate group, the trained ConvNeXt-T (seed 0, 64.1\%, where the three-seed mean is 64.3\%) is fine-tuned on four folds with one fixed recipe, and every image is predicted by a model that never saw its label.
With the human labels of about 7,800 images, top-1 rises by 4.8 points to 68.9\%, \emph{none} recall from 34.1 to 44.8\% and macro-F1 from 60.0 to 66.2.
The same recipe with the model-made labels of the same images lowers top-1 to 62.6\% and \emph{none} recall to 21.0\%, so the gain comes from the labels and not from seeing test-like images: human labels are 6.2 points ahead of model-made ones (95\% interval 5.4 to 7.0).
Training only the linear head on human labels already gives 3.2 of the 4.8 points, so most of what the pseudo-labels cost sits in the decision layer and not in the features.
This is the evidence behind the conclusion of the main paper that cleaner training labels come next.
It is an analysis and not a benchmark row: the models are trained on test images, with one seed and one fold split, and the annotators had seen model suggestions (\cref{sec:supp_anchoring}).

\section{Future work}
\label{sec:supp_future}
The main paper ends on what remains hard: rejecting images outside the 24 classes, the rarest classes, condition, hardware as it leaves service and reading printed identifiers.
Each has a next step that the analyses of this supplement point to.

\paragraph{Cleaner training labels.} In cross-validation on the test set, human labels on about 7,800 images raise top-1 by 4.8 points (\cref{sec:supp_foldft}). Verified labels for a larger part of the training set, starting with the images on which the pseudo-labelers disagree, are the most direct gain.

\paragraph{Rejection.} Training negatives are images the pseudo-labeler read as showing no data-center hardware, while 1,605 of the 2,061 test images annotators marked \emph{none} are look-alikes it had given a class (\cref{sec:supp_foldft}). Out-of-class hardware such as consumer devices, cables and switch line cards belongs in the training negatives.

\paragraph{Synthetic images for rare classes and conditions.} Some classes and conditions are too rare to learn from collected photographs alone. The test set holds 49 SXM GPU modules, 57 GPU server nodes and 133 PDUs, 71 broken and 31 deformed parts, and the training pseudo-labels mark under 4\% of images as other than normal. Image generation conditioned on real photographs of the same classes can add training examples of these: broken, cracked, burnt, bent or crushed parts, stacked and occluded arrangements, and the rare classes in new settings. Generated images would enter training only, flagged in the manifest with the generator and the source image of each, and never validation or test (\cref{sec:supp_datasheet}), so the human-labeled test set measures whether they help. Annotators agree little on broken and deformed ($\alpha$ 0.11 and 0.18, \cref{tab:cond_full}), so such a test needs more doubly answered images of damaged parts first.

\paragraph{Hardware leaving service.} Accuracy falls to 42.1\% on the 178 marketplace and auction photographs (95\% interval about 35 to 49\%), the closest proxy for hardware leaving service (\cref{tab:slices_full}). Photographs taken at a sorting line, as in the envisioned cell of \cref{fig:supp_usecase}, would test the labels where they are meant to be used.

\end{document}